\documentclass[preprint,12pt]{elsarticle}

\usepackage{amssymb}
\usepackage{amssymb}
\usepackage{float}
\usepackage {amsmath}
\usepackage{algorithm}  
\usepackage{algpseudocode}  
\usepackage{amsmath}  
\usepackage{booktabs} 
\usepackage{caption} 
\usepackage{setspace}
\usepackage{subcaption}
\usepackage{hyperref}
\usepackage{threeparttable}
\usepackage{multirow}
\usepackage{doi}
\usepackage{xcolor} 
\usepackage{caption}
\usepackage{mathrsfs}
\usepackage{bm}
\usepackage{adjustbox}

\DeclareMathOperator*{\argmin}{arg\,min} 

\hypersetup{
    colorlinks=true, 
    linkcolor=blue,  
    urlcolor=blue,   
    citecolor=blue   
}

\makeatletter
\renewcommand{\fnum@figure}{\textbf{Fig. \thefigure.}\@gobble}
\makeatother

\makeatletter
\renewcommand{\fnum@table}{\textbf{Table \thetable }\@gobble}
\makeatother

\journal{Elsevier}

\begin{document}

\begin{frontmatter}

\title{
Fault Diagnosis across Heterogeneous Domains 
via Self-Adaptive Temporal-Spatial Attention and Sample Generation
}

\author[a]{Guangqiang Li} 
\ead{guangqiangli@whut.edu.cn}

\author[b]{M. Amine Atoui} 
\ead{amine.atoui@gmail.com}

\author[a]{Xiangshun Li\corref{cor1}} 
\ead{lixiangshun@whut.edu.cn}
\cortext[cor1]{Corresponding author.}

\affiliation[a]{organization={School of Automation},
            addressline={Wuhan University of Technology}, 
            city={Wuhan},
            postcode={430070}, 
            country={PR China}}
\affiliation[b]{organization={The School of Information Technology},
            addressline={Halmstad University}, 
            city={Halmstad},
            country={Sweden}}

\begin{abstract}
Deep learning methods have shown promising performance in fault diagnosis for multimode process. 
Most existing studies assume that the collected health state categories from different operating modes are identical.
However, in real industrial scenarios, these categories typically exhibit only partial overlap.
The incompleteness of the available data and the large distributional differences between the operating modes pose a significant challenge to existing fault diagnosis methods.
To address this problem, a novel fault diagnosis model named self-adaptive temporal-spatial attention network (TSA-SAN) is proposed. 
First, inter-mode mappings are constructed using healthy category data to generate multimode samples.
To enrich the diversity of the fault data, interpolation is performed between healthy and fault samples.
Subsequently, the fault diagnosis model is trained using real and generated data.
The self-adaptive instance normalization is established to suppress irrelevant information while retaining essential statistical features for diagnosis.
In addition, a temporal-spatial attention mechanism is constructed to focus on the key features, thus enhancing the generalization ability of the model. 
The extensive experiments demonstrate that the proposed model significantly outperforms the state-of-the-art methods. 
The code will be available on Github at \href{https://github.com/GuangqiangLi/TSA-SAN}{https://github.com/GuangqiangLi/TSA-SAN}.

\end{abstract}




\end{frontmatter}

\section{Introduction}
\label{sec1}

In industrial manufacturing and production processes, ensuring the safe and reliable operation of the systems has consistently been a central focus for operations and management. 
With the continuous expansion of system scale, the frequency and impact of the faults have correspondingly increased \cite{RN260370, RN246049}. 
Fault diagnosis aims to quickly and accurately infer the fault type, location, magnitude, and occurrence time, thereby providing essential support for predictive maintenance and risk management \cite{RN237009,RN260372, lou2024recent}. 
Recently, benefiting from the application of advanced sensor and communication technologies, plenty of historical monitoring data has been accumulated in the industrial systems.
Consequently, data-driven fault diagnosis methods have attracted considerable attention and have been widely studied.

Deep learning-based fault diagnosis methods have attracted much attention in data-driven methods due to their advantages of automatically extracting features and achieving high diagnostic accuracy.
Various advanced neural network architectures have been extensively applied in this field.
Huang et al. \cite{RN260391} proposed a CNN-LSTM model that combines CNN and LSTM for automatic feature learning. 
Chen et al. \cite{RN265747} developed a fault diagnosis model that utilizes multiscale residual CNN to extract data features across various scales, combines a channel attention mechanism to focus on relevant convolution channels, and leverages the transformer to capture feature dependencies. 
Zhu et al. \cite{RN265748} utilized the multiscale CNN and the bidirectional GRU to learn spatial and temporal features separately. 
Liu et al. \cite{RN265750} employed the multiscale residual network and LSTM network to separately capture multiscale spatial and temporal features. 
Furthermore, Huang et al. \cite{RN265749} integrated the defined outlier-type prior knowledge into CNN layers through a non-parametric attention mechanism. 
This method enables deep integration of prior knowledge with CNNs, thereby achieving high accuracy of fault diagnosis. 

Yet, existing deep learning-based methods typically focus on fault diagnosis under a single operating condition.
Due to the changes in operating environment, production strategy, and load, industrial systems often switch between multiple operating modes. 
The mode switching causes significant differences in data distribution, which may cause samples belonging to the same health state category being dispersed in the feature space, thus posing a challenge for effective classification.
Existing methods employ techniques such as data augmentation, discrepancy metrics, and adversarial learning to address this problem.
Data augmentation-based methods aim to enrich the diversity of training samples, thereby providing a more comprehensive data distribution for the model.
enriching the diversity of the training samples.
Discrepancy-based methods learn common features by minimizing the specific discrepancy criteria between the different domains \cite{RN250833,RN245291}. 
While adversarial learning-based methods learn the domain-invariant feature  through an adversarial game mechanism between the feature extractor and the discriminator \cite{RN245824,RN243235}.
Furthermore, some studies have integrated these techniques to enhance the generalization performance of the model \cite{RN260433, RN260437,RN262328,RN237028}.

The above study assumes \cite{RN250833,RN245291,RN245824,RN243235,RN260433, RN260437,RN262328,RN237028} that data covering all health states are collected under each operating mode. 
However, in practical industrial systems, the high cost and significant risk of fault simulation make it nearly impossible to collect samples of all fault categories under every operating mode\cite{RN265751,RN266412}. 
In more general scenarios, the samples collected under different operating modes typically exhibit less overlap in fault categories. 
Applying traditional methods directly to this scenario faces significant challenges.
First, there exist significant distributional differences across different operating modes, which makes it difficult for data augmentation strategies to adequately cover the data distributions of multiple operating modes.
In addition, aligning only the shared category samples shared across different operating modes may lead to misalignment and confusion among different unshared fault categories, which degrades the diagnostic performance. 
Therefore, constructing a fault diagnosis model from heterogeneous domains is a critical and challenging problem.

To address the above problems, a model named self-adaptive temporal-spatial attention network (TSA-SAN) is presented.
This model mainly contains two processes: sample generation and decision making.
In the sample generation phase, the mapping relationships among different operating modes are established to expand the sample set, thus achieving comprehensive multi-mode coverage.
In addition, interpolation is performed between healthy and fault samples to enrich the distribution of transition fault states.
In the decision-making phase, self-adaptive instance normalization is employed to effectively suppress the interference features associated with different operating modes. 
Meanwhile, an attention mechanism is introduced to focus on critical temporal-spatial regions in the feature maps, so as to extract cross-mode discriminative features and significantly improve the generalization performance of the model.

The main contributions of this paper are as follows.

(1) A novel model named self-adaptive temporal-spatial attention network (TSA-SAN) is proposed to fault diagnosis under heterogeneous domain scenarios.
This model is constructed in domains with class shifts and achieves effective diagnosis across multiple heterogeneous domains.

(2) The distribution alignment sample generation strategy (DASG) and the interpolation-based sample synthesis strategy (ISS) are designed to generate diverse samples. The self-adaptive instance normalization and temporal-spatial attention mechanism are built to remove interference features and extract deep feature representations. 

(3) The extensive experiments demonstrate that the proposed model outperforms the state-of-the-art fault diagnosis models.

The rest of this paper is organized as follows.
\hyperref[sec2]{Section 2} introduces the problem formulation of fault diagnosis across heterogeneous domains and related studies.
\hyperref[sec3]{Section 3} presents the proposed model in detail.
\hyperref[sec4]{Section 4} applies the proposed model to several tasks and thoroughly analyzes the experiment results.
Finally, \hyperref[sec5]{Section 5} gives conclusions.

\section{Problem Formulation }
\label{sec2}

\begin{figure}[!ht]
\centerline{\includegraphics[width=1\columnwidth,height=0.4311\columnwidth]{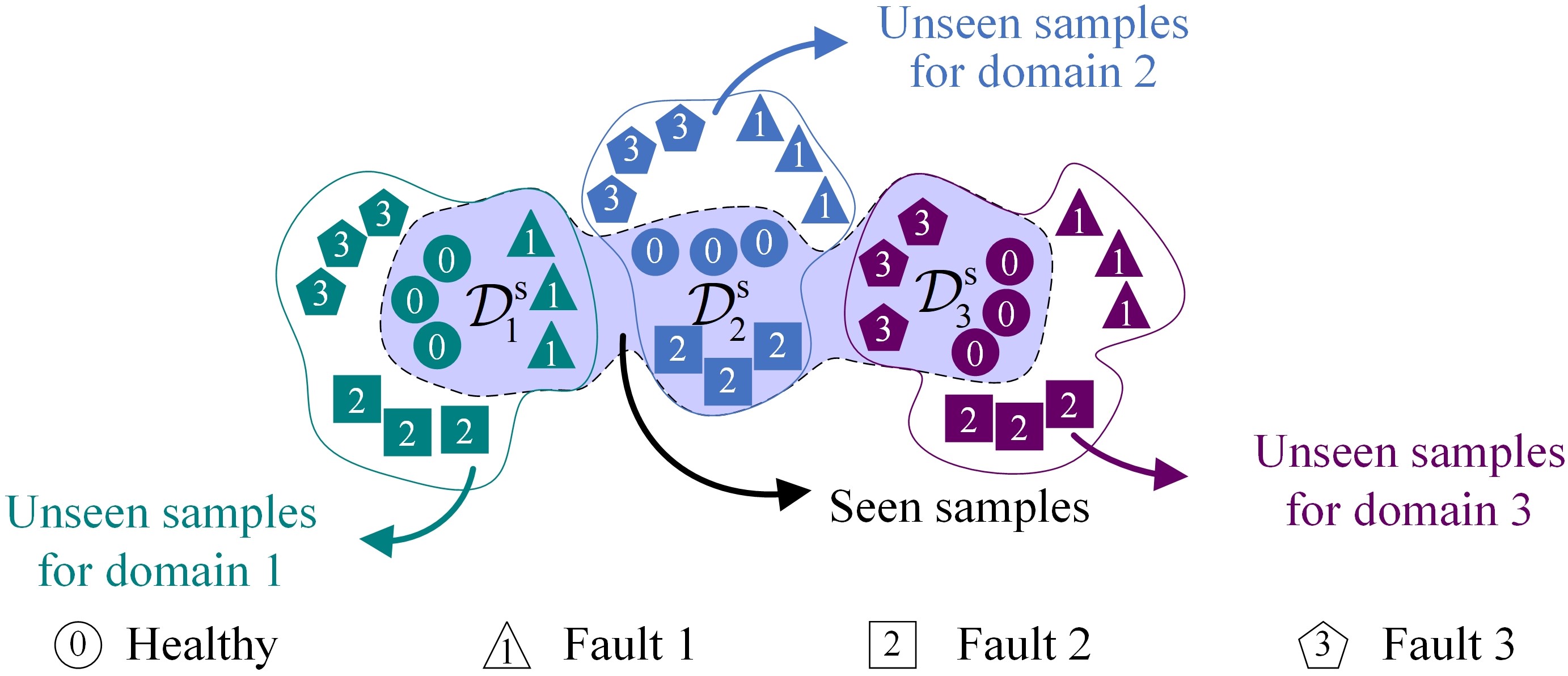}}
	\caption{Visualization of Notations: example with three domains and four health state categories}
	\label{Fig1}
\end{figure}

Let $\mathcal{D}$ denote the sample set for $K$ operating modes (domains), with corresponding health state categories represented by $\mathcal{Y}$. 
$\mathcal{Y}$ contains one healthy category and $\left| \mathcal{Y} \right|-1$ fault categories.
Let $\mathcal{D}^\mathrm{ s }$ denote the collected seen sample set, $\mathcal{D}^\mathrm{ s }=\cup_{i=1}^{K} \{\mathcal{D}_i^\mathrm{ s }\}$.
$\mathcal{D}_i^\mathrm{ s }$ denotes the collected sample set in the $i$-th domain, $\mathcal{D}_i^\mathrm{ s }=\{\bm{x}_{i}^{j}\in\bm{\mathcal{X}}_{i}^{\mathrm{s}},y_{i}^{j}\in\mathcal{Y}_{i}^{\mathrm{s}}\}_{j=1}^{N_{i}^{\mathrm{s}}}$, where $\bm{\mathcal{X}}_{i}^{\mathrm{s}}$ and $\mathcal{Y}_{i}^{\mathrm{s}}$ denote the feature set and the category set of the collected samples in the $i$-th domain. 
$N_{i}^{\mathrm{s}}$ denotes the number of samples collected in the $i$-th domain.
The category set of samples collected from the 
K domains is assumed to satisfy the following constraint: 
$\cup_{i=1}^{K} \mathcal{Y}_i^{\mathrm{s}}=\mathcal{Y}$.
The visualization of these notations is shown in \hyperref[Fig1]{Fig. 1}.
For any two of the $K$ domains, if their collected health state state category sets are identical $\mathcal{Y}_{i}^{\mathrm{s}}=\mathcal{Y}_{j}^{\mathrm{s}}, i<j<K$, they are defined as homogeneous domains; otherwise, they are heterogeneous domains. 

The use of incomplete training datasets, such as $\mathcal{D}^\text{s}$ collected from the $K$ heterogeneous domain, to construct the fault diagnosis model that can be generalized to the complete dataset $\mathcal{D}$ is referred to as fault diagnosis across heterogeneous domains (HDFD).

\begin{figure}[!ht]
\centerline{\includegraphics[width=1\columnwidth,height=0.3397\columnwidth]{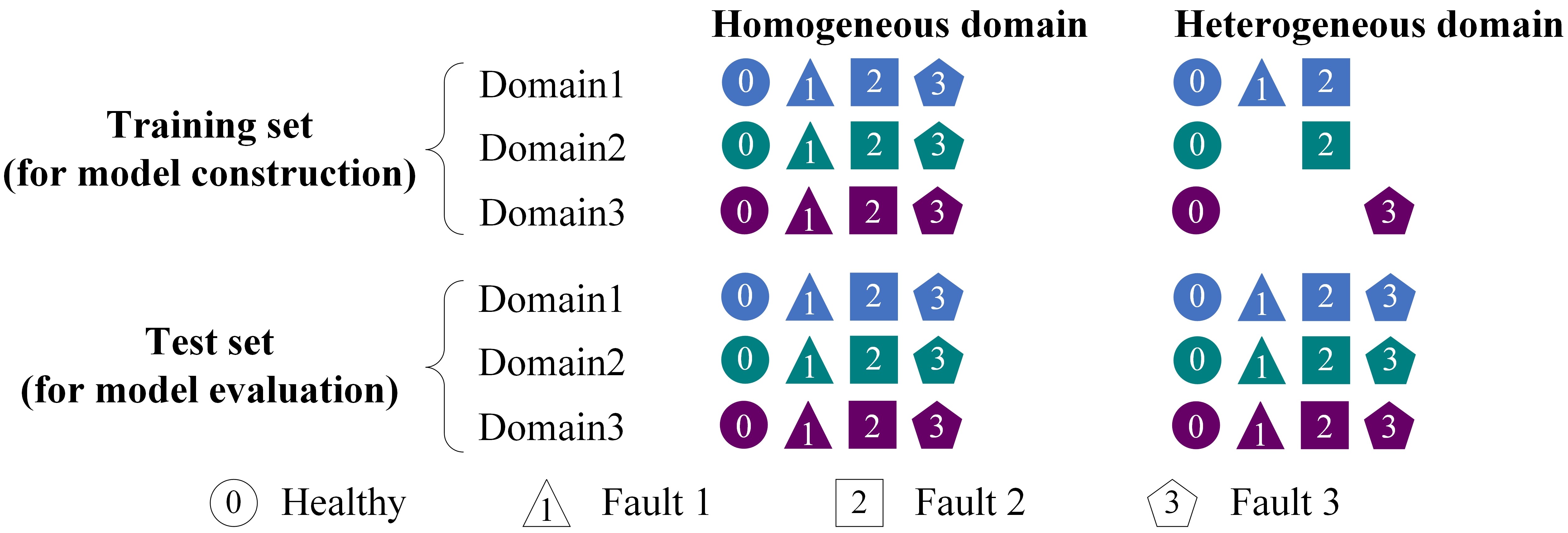}}
	\caption{Illustration of homogeneous vs. heterogeneous domain settings in fault diagnosis.}
	\label{Fig2}
\end{figure}

Fault diagnosis across multiple domains can be categorized into homogeneous-domain and heterogeneous-domain scenarios, depending on whether category shifts exist between domains. 
The differences in their task configurations are illustrated in \hyperref[Fig2]{Figure 2}.
In the homogeneous domain setting, it is generally assumed that the collected health state categories from all domains are the same, i.e., $\mathcal{Y}_{i}^{\mathrm{s}} = \mathcal{Y}$.
Wang et al.\cite{RN260437} developed a generator module to create diverse samples and leveraged adversarial contrastive learning to learn the generalizable feature representations. 
Guo et al. \cite{RN237028} developed a generator to create augmented domains and combined adversarial learning with metric learning to extract common features across domains.
Zhao et al. \cite{RN243183} introduced the Mix-Up to synthesize training samples and effectively captured the feature discriminative features by minimizing the triplet loss function.
Yet, a common limitation of such methods is the assumption of stable category distributions across domains.

In contrast, fault diagnosis across heterogeneous domains (HDFD, see \hyperref[Fig2]{Fig. 2} ) consider a more realistic scenario involving category shifts across domains. 
This setup is frequently encountered in industrial systems, where the health state categories collected from different domains are often inconsistent \cite{RN266421,RN266412}, i.e., $\mathcal{Y}_{i}^{\mathrm{s}} \subseteq \mathcal{Y}$.
Fault diagnosis across heterogeneous domains aims to generalize the fault diagnosis knowledge learned from one domain to other domains. 
As illustrated in \hyperref[Fig2]{Fig. 2}, Fault 1 is exclusively present in Domain 1, Fault 2 occurs in both Domain 1 and Domain 2, and Fault 3 appears only in Domain 3.
The objective of HDFD is to generalize the diagnostic knowledge of Fault 1, learned from Domain 1, to Domain 2 and Domain 3. 
Similarly, the diagnostic knowledge for Fault 2 should be generalized to Domain 3, and that for Fault 3 should be generalized to Domain 1 and Domain 2.

\section{Proposed method}
\label{sec3}
\subsection{Overall Structure}
Although the health state categories in each domain may be incomplete, they can complement each other across domains.
Thus, this study constructs cross-domain mappings using heterogeneous domain data to generate diverse samples. These generated samples, together with real ones, are utilized to train a more generalizable fault diagnosis model.

\begin{figure}[!ht]
\centerline{\includegraphics[width=1.\columnwidth,height=1.124\columnwidth]{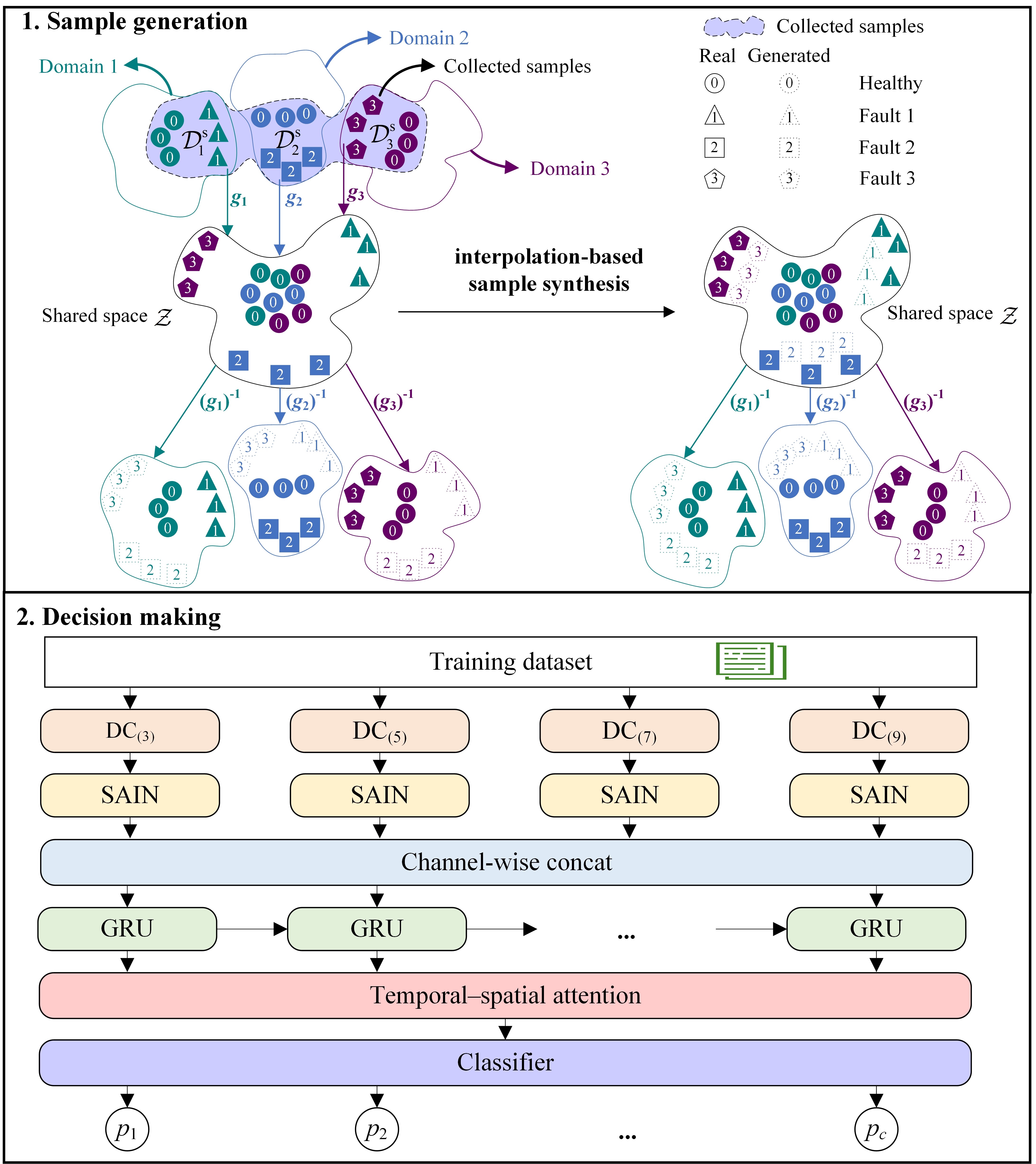}}
	\caption{Construction process of TSA-SAN.}
	\label{Fig3}
\end{figure}

The proposed model consists of two stages: sample generation and decision making. In the sample generation stage, diverse samples are created by mapping relationships of different domains and interpolation of healthy and fault samples. In the decision making stage, temporal-spatial features are adaptively extracted from both real and generated data, enabling accurate classification of the health state category. The construction process the proposed model is shown in \hyperref[Fig3]{Fig. 3}.

\subsection{Sample generation}

In fault diagnosis across heterogeneous domain scenarios, 
the samples collected from each domain cover only a subset of the full set of health state categories, i.e., $\mathcal{Y}_i^{\mathrm{s}} \subseteq \mathcal{Y}, {\forall} \mathcal{D}_i^\mathrm{ s } \subseteq \mathcal{D}^\mathrm{ s }$. 
As a result, the samples collected in all domains constitute an incomplete dataset, i.e., $\mathcal{D}^\mathrm{ s } \subseteq \mathcal{D}$. 
Since deep learning-based fault diagnosis models typically exhibit strong dependence on the distribution of the training data. 
In the absence of prior knowledge and effective data augmentation, these limited samples are insufficient to support the construction of the fault diagnosis model with strong generalization ability.
Thus, it is necessary to generate diverse samples across multiple domains to enhance the completeness of the training dataset.

The sample generation process primarily employs two strategies: distribution alignment sample generation and interpolation-based sample synthesis.
In the distribution alignment sample generation strategy, healthy category samples from different domains are utilized to establish the inter-domain mapping, thus enabling cross-domain sample generation. Specifically, samples from one domain can be input into the learned mapping to generate samples for another domain. 

Consider $\bm{\mathcal{X}}_i^\mathrm{H}, i=1,\dots,K$ denote the healthy category, H, samples from $ K $ domains, while $\bm{\mathcal{X}}_i^{\mathrm{F}_{i}}$
denote the samples whose health state category is exclusive to the $i$ domain.
Assume that there exist $K$ mapping functions $g_i, i=1,\dots,K$ that maps data from the $K$ domains to a shared latent space $\bm{\mathcal{Z}}$. 
In this space, the distributions of samples belonging to the same health state category are approximately aligned. 
These mappings are learned by minimizing the distributional difference of healthy category samples from the $K$ domains,
\begin{equation}
	\begin{aligned}
		\argmin_{g_1,g_2, \dots g_{K}}dist\left(g_1\left(\bm{\mathcal{X}}_1^\mathrm{H}\right),g_2\left(\bm{\mathcal{X}}_2^\mathrm{H},  \right),\dots, g_{K}\left(\bm{\mathcal{X}}_K^\mathrm{H},\right)\right), 
	\end{aligned}
	\label{Eq1}	
\end{equation}
where $dist(.)$ denotes the distance metric function for measuring the distribution difference.
Based on the $K$ mapping functions, the samples whose health state category is exclusive to the $i$-th domain can be mapping to the $j$-th domain. The 
generated fault samples are formulated as,
\begin{equation}
\begin{aligned}
\bm{\tilde{x}}_{j}^{F_{i}} =g_{j}^{-1}\left(g_{i}\left(\bm{x}_{i}^{\mathrm{F}_{i}}\right)\right), \quad \bm{x}_{i}^{\mathrm{F}_{i}} \in \bm{\mathcal{X}}_{i}^{\mathrm{F}_{i}}, 1 \leq i \leq K, 1 \leq j \leq K\\
	\end{aligned}
	\label{Eq2}	
\end{equation}

The learned mappings may fail to accurately align the fault data distributions in the shared latent space, leading to deviations between the generated and actual data distributions. Indeed, there exists a potential risk of extrapolation since the mapping functions are learned solely based on healthy category data.

Given the slow-varying dynamic characteristics commonly observed in process industrial systems, the interpolation sample synthesis strategy are introduced to represent the  gradual transition from healthy state to fault state via Mix-Up between healthy and fault samples. In the shared latent space, the synthesis samples is formulated as,
\begin{equation}
	\begin{aligned}
						\bm{z}_\text{mix}^\text{F}=\lambda \bm{z}^\text{F}+\left( 1-\lambda \right) \bm{z}^\text{N}, \quad \bm{z}^\text{N}\in \bm{\mathcal{Z}}^\text{N}, \bm{z}^\text{F}\in \bm{\mathcal{Z}}^\text{F},
	\end{aligned}
	\label{Eq3}	
\end{equation}
where $\lambda \sim \text{Beta} \left( \alpha,\alpha \right)*0.8+0.2$ and $\alpha=2$, the lower limit of the $\lambda$ is set to 0.2 to prevent synthesized samples with weak fault magnitudes from being overwhelmed by background noise. 
Such low severity samples may have low discriminability and reduce the effectiveness of model training.
The label of each synthesized sample $\bm{z}_\text{mix}^\text{F}$ is assigned to be the same as that of the sample $\bm{z}^\text{F}$.

\subsection{Decision making}
The fault diagnosis model is trained using both collected samples and generated samples, aiming to infer the health state categories by adaptively extracting multi-scale temporal-spatial features. 
This model consists of several modules (see \hyperref[Fig3]{Fig. 3}): multi-scale depthwise convolution (MSDC), self-adaptive instance normalization (SAIN), GRU, temporal-spatial attention mechanism (TSAM) and classifier.

Specifically, the MSDC employs convolution kernels of multiple sizes to extract local features across different receptive fields. The extracted multiscale features are formulated as,
\begin{equation}
	\begin{aligned}
								\bm{f}_\text{MSDC} = \text{DC}_{3} (\bm{x}) \oplus \text{DC}_{5} (\bm{x}) \oplus \text{DC}_{7} (\bm{x}) \oplus \text{DC}_{9} (\bm{x}),
	\end{aligned}
	\label{Eq4}	
\end{equation}
where $\bm{x}$ denotes the input sample, $\text{DC}_{i}$ denotes the depthwise convolution with kernel size $i$, $\oplus$ denotes the channel-wise concatenation.

Instance normalization (IN) has been widely used in cross-domain sample generation tasks due to its ability to suppress domain-specific information. 
In this study, it is utilized to weaken the domain-related components in the feature representations. 
However, since IN normalizes each channel independently and enforces the distribution alignment among different samples, it may inevitably eliminate discriminative information.
To address this limitation, SAIN is proposed to remove domain-specific features while preserving critical discriminative features. 
The normalization process is formulated as follows,

\begin{equation}
	\begin{aligned}
	\bm{f}_\text{SAIN}	& =  \bm{\gamma}_\text{SAIN} \left(\frac{\bm{f}_	\text{MSDC}-\bm{\mu}}{\sqrt{\sigma^{2}+\epsilon}}\right)	+\bm{\beta}_\text{SAIN},\\
    \bm{\gamma}_\text{SAIN} & =h_2 \left( \text{ReLU} \left( h_1 \left( \bm{\mu}    \right)    \right)   \right),\\
    \bm{\beta}_\text{SAIN} & =h_4 \left( \text{ReLU} \left( h_3 \left( \bm{\sigma}    \right)    \right)   \right),
	\end{aligned}
	\label{Eq5}	
\end{equation}
where $\bm{\mu}$ and $\bm{\sigma}$ denote the mean and standard deviation of each channel. 
Unlike IN, which employs fixed scaling and shifting parameters, the proposed SAIN adaptively generates the scaling factor $\gamma$ and shifting factor $\beta$ through four fully connected (FC) layers  $h_i(i=1,2,3,4)$. This design enables the dynamic adjustment of channel-wise feature responses.

The features normalized by the SAIN are subsequently transmitted to GRU to capture long-short-term temporal dependencies. This process is formulated as $\bm{f}_{\text{GRU}}=\text{GRU}\left( \bm{f}_\text{SAIN} \right)$.

\begin{figure}[!ht]
\centerline{\includegraphics[width=1.\columnwidth,height=0.5296\columnwidth]{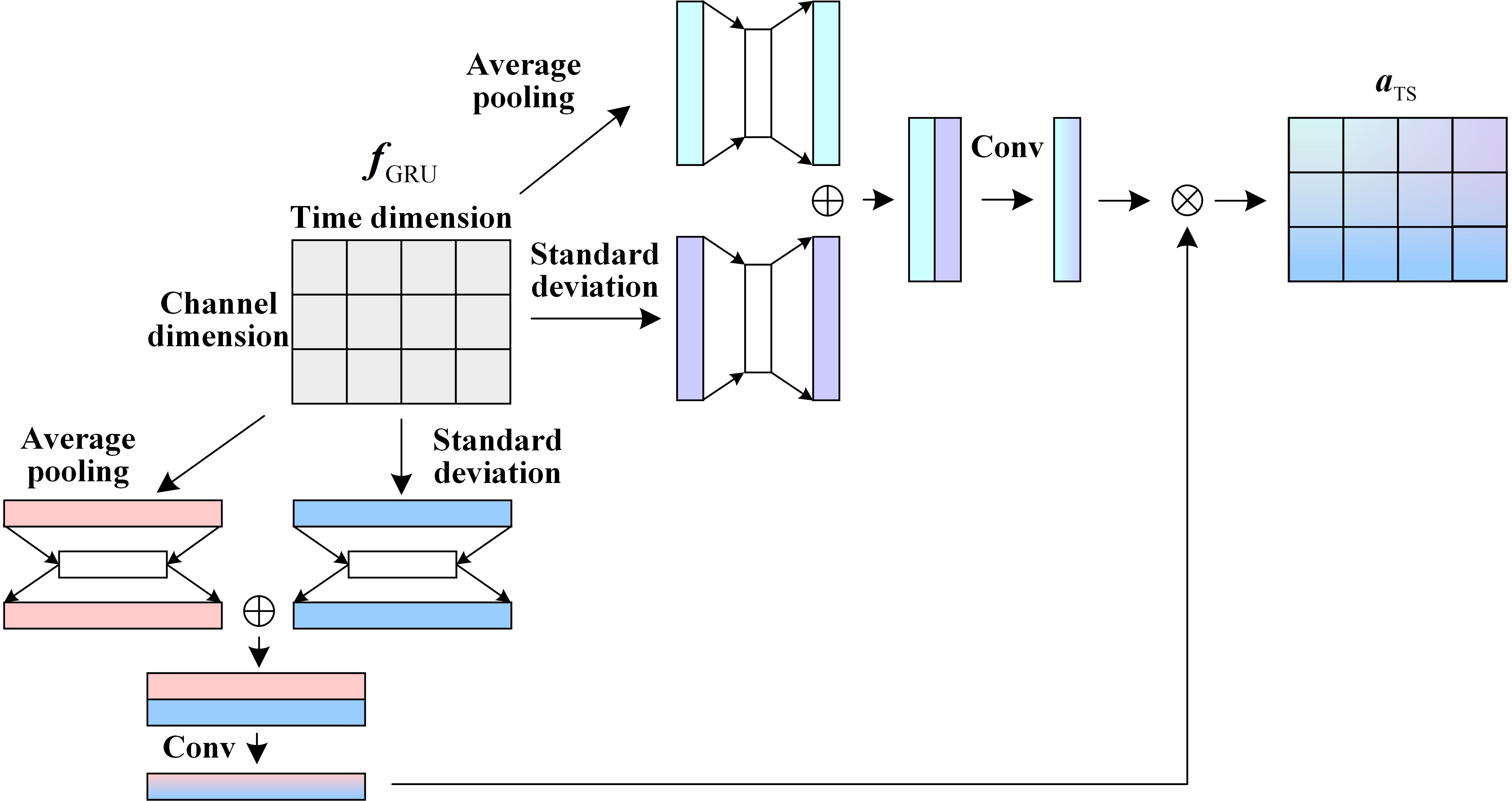}}
	\caption{Structure of TSA.}
	\label{Fig4}
\end{figure}

To emphasize the critical information in the feature maps that is highly relevant to the inference of health state category, a temporal-spatial attention (TSA) mechanism is designed, as shown in \hyperref[Fig4]{Fig. 4}.
The relationships between different timesteps and different channels are jointly used to construct the temporal-spatial attention map, which enable the network to focus on key features in both temporal and channel dimensions.

Specifically, the average pooling (AvgPool) and standard deviation (Std) operations are first performed along the channel and temporal dimensions to obtain global statistical features.
These features are delivered to eight independent FC layers ($h_5-h_{12}$) to generate temporal steady attention map $\bm{a}_{\text{TAP}}$, temporal variability attention map $\bm{a}_{\text{TSD}}$, spatial steady attention map $\bm{a}_{\text{SAP}}$ and spatial variability attention map $\bm{a}_{\text{SSD}}$. 
These operations are formulated as,
\begin{equation}
	\begin{aligned}
			\bm{a}_{\text{TAP}}&=h_{6} \left( \sigma_1 \left( h_{5}(\text{AvgPool}_{\text{C}}\left(\bm{f}_{\text{GRU}}\right))\right)\right), \\
        \bm{a}_{\text{TSD}}&=h_{8} \left( \sigma_1 \left( h_{7}(\text{Std}_{\text{C}}\left(\bm{f}_{\text{GRU}}\right))\right)\right), \\
        			\bm{a}_{\text{SAP}}&=h_{10} \left( \sigma_1 \left( h_{9}(\text{AvgPool}_{\text{T}}\left(\bm{f}_{\text{GRU}}\right))\right)\right), \\
        \bm{a}_{\text{SSD}}&=h_{12} \left( \sigma_1 \left( h_{11}(\text{Std}_{\text{T}}\left(\bm{f}_{\text{GRU}}\right))\right)\right), 
	\end{aligned}
	\label{Eq6}	
\end{equation}
where $\sigma_1$ denotes the ReLU activation function, and the subscripts T and C  denote operations along the temporal and channel dimensions, respectively.

Subsequently, convolution layers are applied to fuse the steady attention map and variability attention map.
The resulting temporal attention map $\bm{a}_\text{T}$ and spatial attention map $\bm{a}_\text{S}$ are formulated as
\begin{equation}
	\begin{aligned}
			\bm{a}_{\text{T}}&=\text{Conv}_\text{T} \left( \bm{a}_{\text{TAP}} \oplus \bm{a}_{\text{TSD}}\right), \\
        \bm{a}_{\text{S}}&=\text{Conv}_\text{S} \left( \bm{a}_{\text{SAP}} \oplus \bm{a}_{\text{SSD}}\right), \\
	\end{aligned}
	\label{Eq7}	
\end{equation}
where $\text{Conv}_\text{S}$ and $\text{Conv}_\text{T}$ denote the convolution layers.
Then $\bm{a}_{\text{T}}$ and $\bm{a}_{\text{S}}$ are fused by matrix product to construct the temporal-spatial attention map $\bm{a}_{\text{TS}}=\bm{a}_{\text{T}}\bm{a}_{\text{S}}$.
Here, $\bm{a}_{\text{TS}}$ represents the importance weights over all temporal-spatial positions in the input feature map $\bm{f}_{\text{GRU}}$. 
$\bm{a}_{\text{TS}}$ is applied to reweight $\bm{f}_{\text{GRU}}$ and enhanced the key components, $\bm{f}_{\text{TS}}=\bm{a}_{\text{TS}} \odot \bm{f}_{\text{GRU}}$, where $\odot$ denotes the Hadamard Product.
The weighted feature maps are summed along the time dimension to obtain the fused feature representation
$\bm{f}_{\text{Fuse}}= {\textstyle \sum_{t}}  \bm{f}_{\text{TS}}$.

Finally, $\bm{f}_{\text{Fuse}}$ is passed through a FC layer followed by a Softmax function to generate the predictive category probability distribution of the input samples $\bm{p}= \text{Softmax} \left( h_{13} \left( \bm{f}_{\text{Fuse}} \right)\right) $, where $h_{13}$ denotes the FC layer.

The cross-entropy loss is employed as the loss function, which is formulated as,
\begin{equation}
	\begin{aligned}
			L=-\sum _{i=1}^N\mathrm{log } \bm{p}_{i,y_i}, \\
	\end{aligned}
	\label{Eq8}	
\end{equation}
where $N$ is the number of samples, $\bm{p}_{i,y_i}$ denotes the probability that the $i$-th sample belongs to its real category $y_i$.
The model parameters are optimized by minimizing this loss using the Adam optimizer.

\section{Experimental study}
\label{sec4}

\subsection{CSTR}
\begin{figure}[!ht]
\centerline{\includegraphics[width=0.6\columnwidth,height=0.2244\columnwidth]{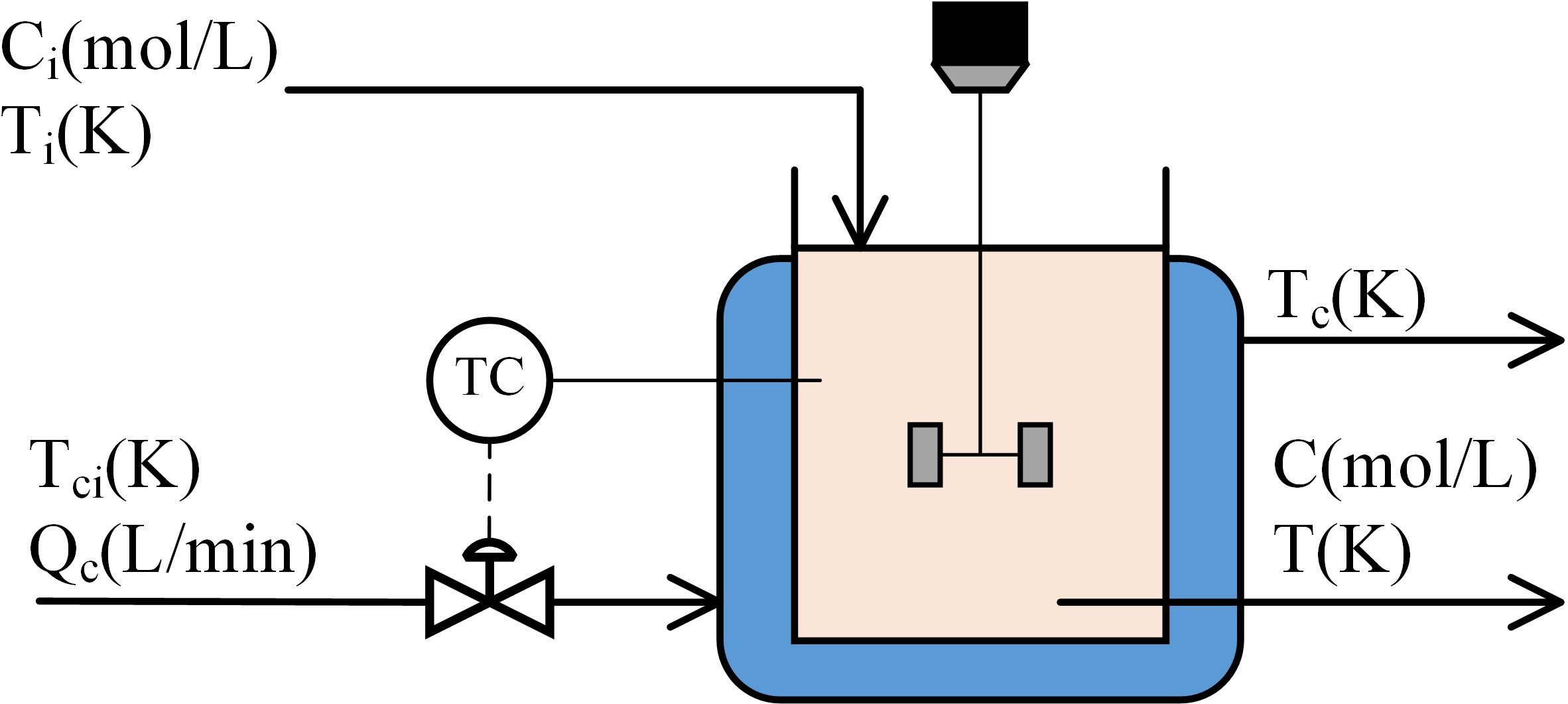}}
	\caption{Structure of closed-loop CSTR \cite{RN237031}.}
	\label{Fig5}
\end{figure}

The closed-loop CSTR \cite{RN237031} is a benchmark platform widely used to assess the effectiveness of fault diagnosis methods.
As shown in \hyperref[Fig5]{Fig. 5}, this system controls the reactor temperature by adjusting the volume flow rate of the cooling water. 
It is described by the following equations,

\begin{equation}
    \begin{aligned}
&\frac{dC}{dt}=\frac{Q}{V}\Big(C_{i}-C\Big)-kC, \\
&\frac{dT}{dt}=\frac{Q}{V}\Big(T_{i}-T\Big)-\frac{(\Delta H_{r})kC}{\rho C_{p}}-\frac{UA}{\rho C_{p}V}\Big(T-T_{c}\Big), \\
&\frac{dT_{c}}{dt}=\frac{Q_{c}}{V_{c}}\Big(T_{ci}-T_{c}\Big)+\frac{UA}{\rho_{c}C_{pc}V_{c}}\Big(T-T_{c}\Big), \\
&k=k_{0}\exp\biggl(\frac{-E}{RT}\biggr),
\end{aligned}
    \label{Eq9}
\end{equation}
where $v_i, (i=1, 2, 3)$ are the process noise, and $k$ is the Arrhenius-type rate constant.
The detailed parameters can be found in \cite{RN237031}. 
The original temperature setpoint was defined as Mode 1 (M1). 
Increasing the setpoint by 5 K and 10 K results in Mode 2 (M2) and Mode 3 (M3), respectively. 
All datasets were generated via the CSTR Simulink model with a total simulation time of 20 hours, and the fault was introduced at the 200th minute. A sliding window of size 64 was applied to process the collected data.

Faults, modes, and tasks used for experimental validation are given in \hyperref[Table1]{Table 1} and \hyperref[Table2]{Table 2}.
For example, in the fourth task, T4, the training dataset for the fault diagnosis model was constructed by integrating five health state categories from M1 and six categories from M2. 
The trained model was then evaluated using the combined dataset consisting of ten health state categories from both M1 and M2.

\begin{table}[!ht]
\centering
	\label{Table1}
		\caption{Faults of CSTR used in HDFD experiments \cite{RN237031}.}
\begin{tabular}{ll}
\hline
\textbf{Health state} & \textbf{Description}           \\ \hline
\textbf{H}                & Healthy                         \\
\textbf{F1}               & $ C_i=C_{i,0}+0.001t $         \\
\textbf{F2}               & $ T_i=T_{i,0}+0.05t $          \\
\textbf{F3}               & $ C=C_0+0.001t $               \\
\textbf{F4}               & $ T=T_0+0.05t $                \\
\textbf{F5}               & $ Q_c=Q_{c,0}-0.1t $           \\
\textbf{F6}               & $ T_{ci}=T_{ci,0}+0.05t $      \\
\textbf{F7}               & $ T_c=T_{c,0}+0.05t $          \\
\textbf{F8}               & $ a=a_0 \text{exp}(-0.0005t) $ \\
\textbf{F9}               & $ b=b_0 \text{exp}(-0.001t) $  \\ \hline
\end{tabular}
\end{table}

\begin{table}[!ht]
\centering
	\label{Table2}
		\caption{Task settings of CSTR used in HDFD experiments.}
\begin{adjustbox}{width=1.2\textwidth,center=\textwidth}
\begin{tabular}{lllllllllllll}
\hline
\textbf{Task}                & \multicolumn{1}{c}{\textbf{Dataset}} & \textbf{Mode} & \textbf{H}   & \textbf{F1}  & \textbf{F2}  & \textbf{F3}  & \textbf{F4}  & \textbf{F5}  & \textbf{F6}  & \textbf{F7}  & \textbf{F8}  & \textbf{F9}  \\ \hline
\multirow{3}{*}{\textbf{T1}} & \multirow{2}{*}{Training}            & M1            & $\checkmark$ & $\checkmark$ & $\checkmark$ & $\checkmark$ & $\checkmark$ & $\checkmark$ & $\checkmark$ & $\checkmark$ & $\checkmark$ &              \\\cline{3-13} 
                             &                                      & M2            & $\checkmark$ & $\checkmark$ & $\checkmark$ & $\checkmark$ & $\checkmark$ & $\checkmark$ & $\checkmark$ & $\checkmark$ &              & $\checkmark$ \\ \cline{2-13} 
                             & Test                                 & M1, M2        & $\checkmark$ & $\checkmark$ & $\checkmark$ & $\checkmark$ & $\checkmark$ & $\checkmark$ & $\checkmark$ & $\checkmark$ & $\checkmark$ & $\checkmark$ \\ \hline\hline
\multirow{3}{*}{\textbf{T2}} & \multirow{2}{*}{Training}            & M1            & $\checkmark$ & $\checkmark$ & $\checkmark$ & $\checkmark$ & $\checkmark$ & $\checkmark$ & $\checkmark$ & $\checkmark$ &              & $\checkmark$ \\\cline{3-13} 
                             &                                      & M3            & $\checkmark$ &              &              & $\checkmark$ & $\checkmark$ &              & $\checkmark$ & $\checkmark$ & $\checkmark$ &               \\ \cline{2-13} 
                             & Test                                 & M1, M3        & $\checkmark$ & $\checkmark$ & $\checkmark$ & $\checkmark$ & $\checkmark$ & $\checkmark$ & $\checkmark$ & $\checkmark$ & $\checkmark$ & $\checkmark$ \\ \hline\hline
\multirow{3}{*}{\textbf{T3}} & \multirow{2}{*}{Training}            & M2            & $\checkmark$ & $\checkmark$ &              & $\checkmark$ &              &              &              &              &              & $\checkmark$ \\\cline{3-13} 
                             &                                      & M3            & $\checkmark$ & $\checkmark$ & $\checkmark$ &              & $\checkmark$ & $\checkmark$ & $\checkmark$ & $\checkmark$ & $\checkmark$ &                                      \\\cline{2-13} 
                             & Test                                 & M2, M3        & $\checkmark$ & $\checkmark$ & $\checkmark$ & $\checkmark$ & $\checkmark$ & $\checkmark$ & $\checkmark$ & $\checkmark$ & $\checkmark$ & $\checkmark$ \\ \hline\hline
\multirow{3}{*}{\textbf{T4}} & \multirow{2}{*}{Training}            & M1            & $\checkmark$ & $\checkmark$ & $\checkmark$ & $\checkmark$ & $\checkmark$ &              &              &              &              &              \\\cline{3-13} 
                             &                                      & M2            & $\checkmark$ &              &              &              &              & $\checkmark$ & $\checkmark$ & $\checkmark$ & $\checkmark$ & $\checkmark$ \\ \cline{2-13} 
                             & Test                                 & M1, M2        & $\checkmark$ & $\checkmark$ & $\checkmark$ & $\checkmark$ & $\checkmark$ & $\checkmark$ & $\checkmark$ & $\checkmark$ & $\checkmark$ & $\checkmark$ \\ \hline\hline
\multirow{3}{*}{\textbf{T5}} & \multirow{2}{*}{Training}            & M1            & $\checkmark$ & $\checkmark$ & $\checkmark$ & $\checkmark$ & $\checkmark$ &              &              &              &              &              \\\cline{3-13} 
                             &                                      & M3            & $\checkmark$ &              &              &              &              & $\checkmark$ & $\checkmark$ & $\checkmark$ & $\checkmark$ & $\checkmark$ \\ \cline{2-13} 
                             & Test                                 & M1, M3        & $\checkmark$ & $\checkmark$ & $\checkmark$ & $\checkmark$ & $\checkmark$ & $\checkmark$ & $\checkmark$ & $\checkmark$ & $\checkmark$ & $\checkmark$ \\ \hline\hline
\multirow{3}{*}{\textbf{T6}} & \multirow{2}{*}{Training}            & M2            & $\checkmark$ & $\checkmark$ & $\checkmark$ & $\checkmark$ & $\checkmark$ &              &              &              &              &              \\\cline{3-13} 
                             &                                      & M3            & $\checkmark$ &              &              &              &              & $\checkmark$ & $\checkmark$ & $\checkmark$ & $\checkmark$ & $\checkmark$ \\\cline{2-13} 
                             & Test                                 & M2, M3        & $\checkmark$ & $\checkmark$ & $\checkmark$ & $\checkmark$ & $\checkmark$ & $\checkmark$ & $\checkmark$ & $\checkmark$ & $\checkmark$ & $\checkmark$ \\ \hline\hline
\multirow{4}{*}{\textbf{T7}} & \multirow{3}{*}{Training}            & M1            & $\checkmark$ & $\checkmark$ & $\checkmark$ & $\checkmark$ & $\checkmark$ &              &              &              &              &              \\\cline{3-13} 
                             &                                      & M2            & $\checkmark$ &              &              &              &              & $\checkmark$ & $\checkmark$ & $\checkmark$ & $\checkmark$ & $\checkmark$ \\\cline{3-13} 
                             &                                      & M3            & $\checkmark$ &              &              &              &              &              &              &              &              &              \\\cline{2-13} 
                             & Test                                 & M1, M2, M3    & $\checkmark$ & $\checkmark$ & $\checkmark$ & $\checkmark$ & $\checkmark$ & $\checkmark$ & $\checkmark$ & $\checkmark$ & $\checkmark$ & $\checkmark$ \\ \hline\hline
\multirow{4}{*}{\textbf{T8}} & \multirow{3}{*}{Training}            & M1            & $\checkmark$ & $\checkmark$ & $\checkmark$ & $\checkmark$ & $\checkmark$ &              &              &              &              &              \\\cline{3-13} 
                             &                                      & M3            & $\checkmark$ &              &              &              &              & $\checkmark$ & $\checkmark$ & $\checkmark$ & $\checkmark$ & $\checkmark$ \\\cline{3-13} 
                             &                                      & M2            & $\checkmark$ &              &              &              &              &              &              &              &              &              \\\cline{2-13} 
                             & Test                                 & M1, M2, M3    & $\checkmark$ & $\checkmark$ & $\checkmark$ & $\checkmark$ & $\checkmark$ & $\checkmark$ & $\checkmark$ & $\checkmark$ & $\checkmark$ & $\checkmark$ \\ \hline\hline
\multirow{4}{*}{\textbf{T9}} & \multirow{3}{*}{Training}            & M2            & $\checkmark$ & $\checkmark$ & $\checkmark$ & $\checkmark$ & $\checkmark$ &              &              &              &              &              \\\cline{3-13} 
                             &                                      & M3            & $\checkmark$ &              &              &              &              & $\checkmark$ & $\checkmark$ & $\checkmark$ & $\checkmark$ & $\checkmark$ \\ \cline{3-13} 
                             &                                      & M1            & $\checkmark$ &              &              &              &              &              &              &              &              &              \\\cline{2-13} 
                             & Test                                 & M1, M2, M3    & $\checkmark$ & $\checkmark$ & $\checkmark$ & $\checkmark$ & $\checkmark$ & $\checkmark$ & $\checkmark$ & $\checkmark$ & $\checkmark$ & $\checkmark$ \\ \hline
\end{tabular}
\end{adjustbox}
\end{table}

\subsection{TE process}
\begin{figure}[!ht]
\centerline{\includegraphics[width=1.\columnwidth,height=0.5296\columnwidth]{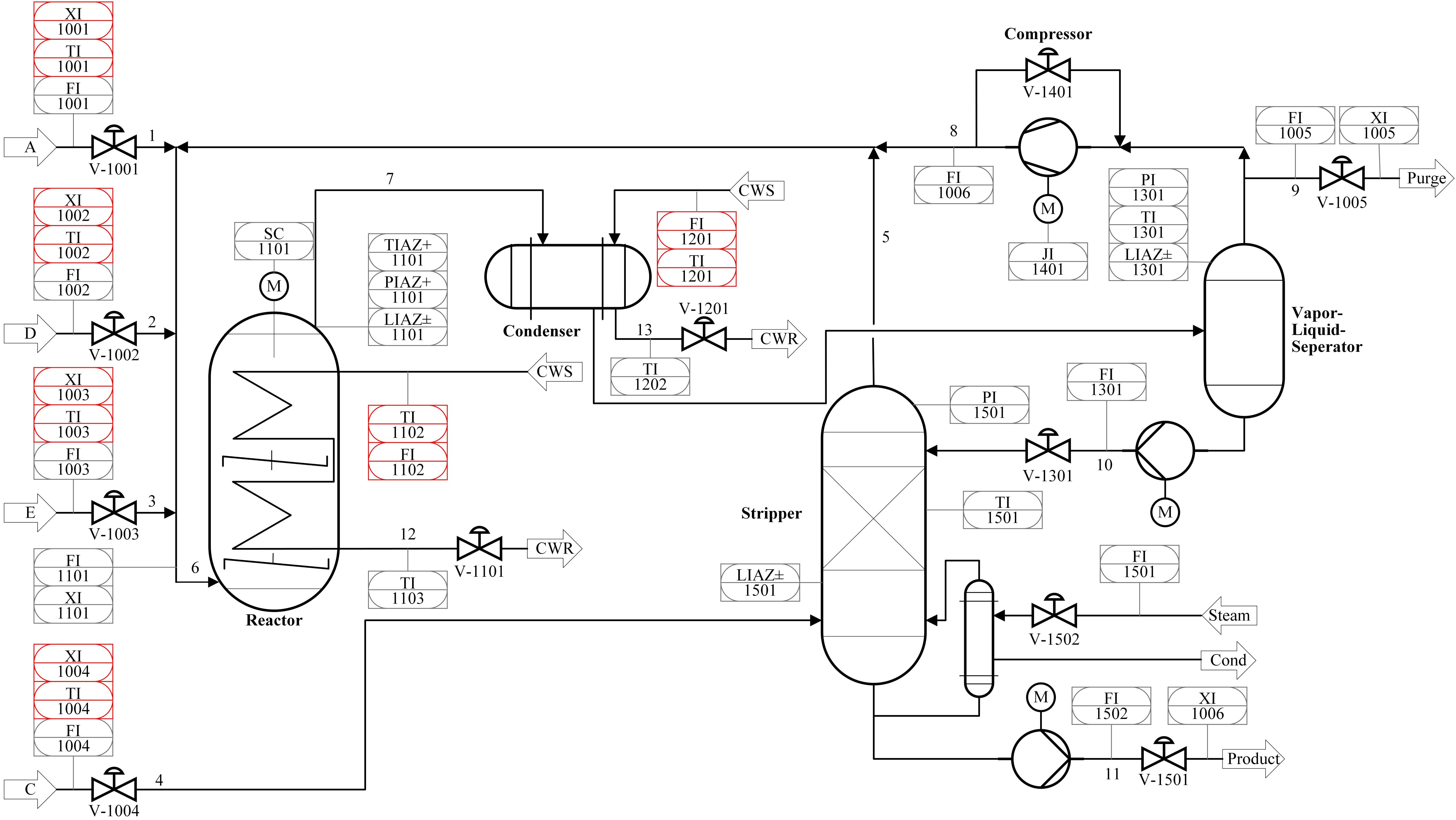}}
	\caption{P\&ID of the revised process model \cite{RN237021}.}
	\label{Fig6}
\end{figure}

The Tennessee Eastman (TE) process \cite{RN237020} is also widely employed in the evaluation of fault diagnostic model performance.
Its structure is shown in \hyperref[Fig6]{Fig. 6}.
Bathelt et al. \cite{RN237021} implemented the TE process using SIMULINK, while Liu et al. \cite{RN260410} implemented the simulation for different operating modes. 
It was simulated with 16 health state categories across 6 operating modes in this study, see \hyperref[Table3]{Table 3} and  \hyperref[Table4]{Table 4}. All experimental datasets were generated based on the TE simulation platform, where the sampling time was 3 minutes and the simulation time was 100 hours, with faults injected at the 30th hour. 
The same sliding window processing used in the CSTR was employed. 
The details settings of the four different 
diagnostic tasks considered in HDFD experimental validation are provided in \hyperref[Table5]{Table 5}.

\begin{table}[!ht]
\centering
	\label{Table3}
		\caption{Faults of TE process used in HDFD experiements\cite{RN237021}.}
\begin{tabular}{lll}
\hline
\textbf{\begin{tabular}[c]{@{}l@{}}Health \\ state\end{tabular}} & \textbf{Description}                                     & \textbf{Type}    \\ \hline
\textbf{H}                                                           & Healthy                                                   & -                \\
\textbf{F1}                                                          & A/C feed ratio, B composition constant (stream 4)        & Step             \\
\textbf{F2}                                                          & B composition, A/C ratio constant (Stream 4)             & Step             \\
\textbf{F4}                                                          & Reactor cooling water inlet temperature                  & Step             \\
\textbf{F6}                                                          & A feed loss (stream 1)                                   & Step             \\
\textbf{F7}                                                          & C header pressure loss - reduced availability (stream 4) & Step             \\
\textbf{F8}                                                          & A, B, C feed composition (stream 4)                      & Random variation \\
\textbf{F10}                                                         & C feed temperature (stream 4)                            & Random variation \\
\textbf{F11}                                                         & Reactor cooling water inlet temperature                  & Random variation \\
\textbf{F12}                                                         & Condenser cooling water inlet temperature                & Random variation \\
\textbf{F13}                                                         & Reaction kinetics                                        & Drift            \\
\textbf{F14}                                                         & Reactor cooling water valve                              & Sticking         \\
\textbf{F17-20}                                                      & Unknown                                                  & Unknown          \\ \hline
\end{tabular}
\end{table}

\begin{table}[!ht]
\centering
	\label{Table4}
		\caption{Modes of TE process used in HDFD experiements\cite{RN237021}.}
\begin{tabular}{ll}
\hline
\textbf{Mode} & \textit{G/H} \textbf{mass ratio and production rate} \\ \hline
\textbf{M1}       & 50/50, \textit{G}: 7038kg/h, \textit{H}: 7038kg/h             \\
\textbf{M2}       & 10/90, \textit{G}: 1408kg/h, \textit{H}: 12669kg/h            \\
\textbf{M3}       & 90/10, \textit{G}: 10000kg/h, \textit{H}: 1111kg/h            \\
\textbf{M4}       & 50/50, maximum production rate              \\
\textbf{M5}       & 10/90, maximum production rate              \\
\textbf{M6}       & 90/10, maximum production rate              \\ \hline
\end{tabular}
\end{table}

\begin{table}[!ht]
\centering
	\label{Table5}
	\caption{Task settings of TE process used in HDFD experiments.}
\begin{adjustbox}{width=1.2\textwidth,center=\textwidth}
\begin{tabular}{lllllllllll}
\hline
\textbf{Task}                & \multicolumn{1}{c}{\textbf{Dataset}} & \textbf{Mode} & \textbf{H}   & \textbf{F1}  & \textbf{F2}  & \textbf{F4}  & \textbf{F6-F8} & \textbf{F10} & \textbf{F11-F14} & \textbf{F17-F20} \\ \hline
\multirow{3}{*}{\textbf{T1}} & \multirow{2}{*}{Training}            & M1            & $\checkmark$ & $\checkmark$ & $\checkmark$ & $\checkmark$ & $\checkmark$   & $\checkmark$ &                  &                  \\ \cline{3-11} 
                             &                                      & M4            & $\checkmark$ &              &              &              &                &              & $\checkmark$     & $\checkmark$     \\\cline{2-11} 
                             & Test                                 & M1, M4        & $\checkmark$ & $\checkmark$ & $\checkmark$ & $\checkmark$ & $\checkmark$   & $\checkmark$ & $\checkmark$     & $\checkmark$     \\  \hline\hline
\multirow{3}{*}{\textbf{T2}} & \multirow{2}{*}{Training}            & M2            & $\checkmark$ & $\checkmark$ & $\checkmark$ & $\checkmark$ & $\checkmark$   & $\checkmark$ &                  &                  \\\cline{3-11} 
                             &                                      & M5            & $\checkmark$ &              &              &              &                &              & $\checkmark$     & $\checkmark$     \\ \cline{2-11} 
                             & Test                                 & M2, M5        & $\checkmark$ & $\checkmark$ & $\checkmark$ & $\checkmark$ & $\checkmark$   & $\checkmark$ & $\checkmark$     & $\checkmark$     \\ \hline\hline
\multirow{3}{*}{\textbf{T3}} & \multirow{2}{*}{Training}            & M3            & $\checkmark$ & $\checkmark$ & $\checkmark$ & $\checkmark$ & $\checkmark$   & $\checkmark$ &                  &                  \\\cline{3-11} 
                             &                                      & M6            & $\checkmark$ &              &              &              &                &              & $\checkmark$     & $\checkmark$     \\ \cline{2-11} 
                             & Test                                 & M3, M6        & $\checkmark$ & $\checkmark$ & $\checkmark$ & $\checkmark$ & $\checkmark$   & $\checkmark$ & $\checkmark$     & $\checkmark$     \\ \hline\hline
\multirow{3}{*}{\textbf{T4}} & \multirow{2}{*}{Training}            & M1            & $\checkmark$ & $\checkmark$ & $\checkmark$ & $\checkmark$ & $\checkmark$   & $\checkmark$ &                  &                  \\ \cline{3-11} 
                             &                                      & M2            & $\checkmark$ &              &              &              &                &              & $\checkmark$     & $\checkmark$     \\ \cline{2-11} 
                             & Test                                 & M1, M2        & $\checkmark$ & $\checkmark$ & $\checkmark$ & $\checkmark$ & $\checkmark$   & $\checkmark$ & $\checkmark$     & $\checkmark$     \\ \hline
\end{tabular}\end{adjustbox}
\end{table}

\subsection{Model Implementation details}
This section first determines the cross-domain mapping model and then gives the parameters of the fault diagnosis model.  
Each stable operating mode in a multimode process is formulated as \cite{multimode2021},
\begin{equation}
	\left|\frac{x(t)-x\left(t_{0}\right)}{t-t_{0}}\right|<T_{x} \quad \forall t \in\left[t_{0}, t_{0}+\triangle t\right], 
	\label{Eq10}	
\end{equation} 
where $x\left(t\right)$ denotes the value of the variable at time $t$, and $ T_{x}$ is a threshold.
This formulation indicates that the healthy category data in each domain exhibits strong stability.
Directly learning inter-domain mappings through neural networks is easy to overfit the healthy category samples, which limits cross-domain generalization. 
Therefore, it is assumed that the sample distributions in different domains can be approximately aligned by the mean and standard deviation of their respective healthy category samples.

The detailed architecture of the proposed model is summarized in  \hyperref[Table6]{Table 6}, where $v$ is the number of monitoring variables. 
The training parameters are configured as follows: batch size = 512, learning rate = $0.01\times0.3^{\left(epoch//3\right)}$, and the number of training epochs = 30.

\begin{table}[!ht]
	\centering
	\caption{Detailed architecture of TSA-SAN.}
	\label{Table6}
	\begin{tabular}{lll}
\hline
\textbf{Symbols}               & \textbf{Input, output}                & \textbf{Detailed architecture}                                                                                                                                              \\ \hline
$\text{DC}_{\left( 3 \right)}$ & ($v\times 64$),($v\times$64)          & Depthwise convolution, kernel=3                                                                                                                                             \\
$\text{DC}_{\left( 5 \right)}$ & ($v\times 64$),($v\times$64)          & Depthwise convolution, kernel=5                                                                                                                                             \\
$\text{DC}_{\left( 7 \right)}$ & ($v\times 64$),($v\times$64)          & Depthwise convolution, kernel=7                                                                                                                                             \\
$\text{DC}_{\left( 9 \right)}$ & ($v\times 64$),($v\times$64)          & Depthwise convolution, kernel=9                                                                                                                                             \\
SAIN                           & ($v\times 64$),($v\times$64)          & Fully connected layer, ($v \rightarrow v \rightarrow v$) $\times 2$                                                                                                         \\
GRU                            & ($4v\times 64$),($2v\times$64)        & -                                                                                                                                                                           \\
\multirow{2}{*}{TSAM}          & ($2v\times 64$),($2v\times 64$)       & \begin{tabular}[c]{@{}l@{}}Fully connected layer, ($v \rightarrow (v//16) \rightarrow v$) $\times 2$, \\ ($64 \rightarrow (64//16) \rightarrow 64$) $\times 2$\end{tabular} \\
                               & ($2v\times 64$),($2v$)                & Convolution layer $\times 2$, kernel=1                                                                                                                                      \\
Classifier                     & ($2v$),($\left| \mathcal{Y} \right|$) & Fully connected layer, ($2v \rightarrow \left| \mathcal{Y} \right| $)                                                                                                       \\ \hline
\end{tabular}
\end{table}

\subsection{Comparative models and Evaluation metrics}
\subsubsection{Models}
To validate the effectiveness of the proposed model, several state-of-the-art fault diagnosis methods were selected for comparison:
\begin{itemize}
\item CNN-LSTM \cite{RN260391}: Combines convolutional neural networks and long short-term memory networks to extract both spatial and temporal features. It performs well in complex fault diagnosis tasks.
\item MCNN-DBiGRU \cite{RN265748}: Uses a multi-scale convolutional network to extract features at different scales and a bidirectional gated recurrent unit to capture time-based patterns in both directions.
\item MSG-ACN \cite{RN260437}: Generates diverse data samples using multi-scale style generation and applies contrastive learning to improve the model’s ability to generalize to new data.
\item MGAMN \cite{RN237028}: Creates domain-extended samples and combines adversarial and metric learning to extract features that are effective across different conditions.
\item SDAGN \cite{RN243183}: Synthesizes new samples and uses triplet loss to learn consistent, generalizable features for fault diagnosis.
\end{itemize}

Among these methods, CNN-LSTM \cite{RN260391} and MCNN-DBiGRU \cite{RN265748} have shown strong performance in single-domain fault diagnosis. In contrast, MSG-ACN \cite{RN260437}, MGAMN \cite{RN237028}, and SDAGN \cite{RN243183} are designed for fault diagnosis in homogeneous domains.
They are employed to evaluate the challenges of fault diagnosis across heterogeneous domains in multimode processes.

\subsubsection{Evaluation metrics}
Accuracy (ACC), fault diagnosis rate (FDR) and false positive rate (FPR) \cite{RN260387} are adopted as evaluation metrics and they are defined as follows,

\begin{equation}
    ACC=\frac{\sum_{l=1}^LTP_l}{\sum_{l=1}^LTP_l+\sum_{l=1}^LFN_l},
    \label{Eq11}
\end{equation}

\begin{equation}
    FDR_l=\frac{TP_l}{TP_l+FN_l},
    \label{Eq12}
\end{equation}

\begin{equation}
    FPR_l=\frac{FP_l}{FP_l+TN_l},
    \label{Eq13}
\end{equation}
where $TP$, $TN$, $FP$, and $FN$ denote the numbers of true positive, true negative, false positive, and false negative samples, respectively.
Accuracy measures the overall capability of the model to infer the health states. 
FDR indicates the proportion of samples labeled $l$ that are correctly classified as $l$. 
FPR represents the proportion of samples not labeled $l$ that are incorrectly classified as $l$. 
The model exhibits better performance when the ACC and FDR are higher and the FPR is lower.

\subsection{Experiment results and analysis}

\subsubsection{CSTR}
The accuracy comparison of different models on the CSTR dataset is shown in \hyperref[Table7]{Table 7}.
The proposed TSA-SAN model obtained the highest diagnostic accuracy on eight tasks, and ranked second on task T1, with only a 0.04\% gap from  SDAGN. 
Moreover, it attained the highest average accuracy, which is 14.47\%, 13.62\%, 15.75\%, 15.06\% and 8.60\% higher than CNN-LSTM, MCNN-DBiGRU, MSG-ACN, MGAMN and SDAGN, respectively.
These results strongly demonstrate the superior generalization ability of the proposed model under fault diagnosis across heterogeneous domain scenarios.

\begin{table}[!ht]
	\centering
	\caption{Accuracy results on CSTR dataset.}
	\label{Table7}
      \begin{adjustbox}{width=1.4\textwidth,center=\textwidth}
\begin{tabular}{lllllllllll}
\hline
\textbf{Model}       & \textbf{T1}       & \textbf{T2}      & \textbf{T3}      & \textbf{T4}      & \textbf{T5}      & \textbf{T6}      & \textbf{T7}      & \textbf{T8}      & \textbf{T9}      & \textbf{Avg}     \\ \hline
\textbf{CNN-LSTM}    & 78.68\%           & 72.85\%          & 81.70\%          & 87.39\%          & 75.71\%          & 92.43\%          & 88.32\%          & 84.67\%          & 83.06\%          & 82.76\%          \\
\textbf{MCNN-DBiGRU} & 85.22\%           & 68.41\%          & 69.95\%          & 91.65\%          & 79.37\%          & 95.36\%          & 93.10\%          & 85.17\%          & 84.27\%          & 83.61\%          \\
\textbf{MSG-ACN}     & 76.97\%           & 69.23\%          & 82.21\%          & 93.65\%          & 75.79\%          & 85.43\%          & 94.05\%          & 75.76\%          & 80.22\%          & 81.48\%          \\
\textbf{MGAMN}       & 78.11\%           & 74.64\%          & 79.78\%          & 90.84\%          & 79.47\%          & 84.04\%          & 92.99\%          & 79.83\%          & 79.86\%          & 82.17\%          \\
\textbf{SDAGN}       & 95.53\%           & 67.74\%          & 93.99\%          & \textbf{98.37\%} & 84.37\%          & 85.59\%          & 97.55\%          & 89.62\%          & 84.91\%          & 88.63\%          \\
\textbf{TSA-SAN}     & \textbf{100.00\%} & \textbf{96.64\%} & \textbf{94.98\%} & 98.33\%          & \textbf{95.43\%} & \textbf{99.14\%} & \textbf{97.62\%} & \textbf{96.14\%} & \textbf{96.81\%} & \textbf{97.23\%} \\ \hline
\end{tabular}

\end{adjustbox}
\end{table}

Using task T6 as an example, the FDR and FPR scores of different models are summarized in \hyperref[Table8]{Table 8}. 
The proposed TSA-SAN model achieved the highest average FDR and the lowest average FPR among all models. 
In addition, it attained a minimum FDR of 0.9576 and a maximum FPR of only 0.0052, which are significantly better than the comparison model. These results show the high performance of the proposed TSA-SAN model.

\begin{table}[!ht]
	\caption{FDR and FPR results on CSTR dataset.}
	\label{Table8}
    \begin{adjustbox}{width=1.4\textwidth,center=\textwidth}
   \begin{tabular}{lllllllllllll}
\hline
\multirow{2}{*}{\textbf{\begin{tabular}[c]{@{}l@{}}Fault \\ No.\end{tabular}}} & \multicolumn{2}{c}{\textbf{CNN-LSTM}} & \multicolumn{2}{l}{\textbf{MCNN-DBiGRU}} & \multicolumn{2}{c}{\textbf{MSG-ACN}} & \multicolumn{2}{c}{\textbf{MGAMN}} & \multicolumn{2}{c}{\textbf{SDAGN}} & \multicolumn{2}{c}{\textbf{TSA-SAN}} \\ \cline{2-13} 
                                                                               & \textbf{FDR}      & \textbf{FPR}      & \textbf{FDR}        & \textbf{FPR}       & \textbf{FDR}      & \textbf{FPR}     & \textbf{FDR}     & \textbf{FPR}    & \textbf{FDR}     & \textbf{FPR}    & \textbf{FDR}      & \textbf{FPR}     \\ \hline
\textbf{N}                                                                     & 1.0000            & \textit{0.0315}   & 1.0000              & 0.0072             & 0.9000            & 0.0318           & 1.0000           & \textit{0.1295} & 1.0000           & \textit{0.1448} & 1.0000            & 0.0000           \\
\textbf{F1}                                                                    & 1.0000            & 0.0000            & 1.0000              & 0.0000             & 0.8167            & 0.0000           & 0.8258           & 0.0024          & 0.9727           & 0.0000          & 1.0000            & 0.0041           \\
\textbf{F2}                                                                    & 1.0000            & 0.0000            & 1.0000              & 0.0059             & 1.0000            & 0.0000           & 1.0000           & 0.0004          & 1.0000           & 0.0000          & 1.0000            & 0.0000           \\
\textbf{F3}                                                                    & 1.0000            & 0.0000            & 1.0000              & 0.0031             & 1.0000            & 0.0000           & 1.0000           & 0.0011          & 1.0000           & 0.0000          & 1.0000            & 0.0000           \\
\textbf{F4}                                                                    & 0.9455            & 0.0254            & 0.9258              & 0.0000             & 1.0000            & \textit{0.1050}  & 0.8742           & 0.0000          & 0.9409           & 0.0000          & 0.9970            & \textit{0.0052}  \\
\textbf{F5}                                                                    & \textit{0.5667}   & 0.0000            & \textit{0.7985}     & 0.0000             & 1.0000            & 0.0226           & 0.8758           & 0.0213          & 1.0000           & 0.0007          & 0.9667            & 0.0000           \\
\textbf{F6}                                                                    & 1.0000            & 0.0000            & 1.0000              & 0.0000             & 1.0000            & 0.0000           & 1.0000           & 0.0000          & 1.0000           & 0.0000          & 1.0000            & 0.0000           \\
\textbf{F7}                                                                    & 1.0000            & 0.0000            & 0.9152              & 0.0080             & 1.0000            & 0.0000           & 0.9697           & 0.0000          & 1.0000           & 0.0000          & 1.0000            & 0.0000           \\
\textbf{F8}                                                                    & 0.7924            & 0.0250            & 0.9348              & \textit{0.0270}    & \textit{0.1500}   & 0.0009           & \textit{0.1530}  & 0.0020          & \textit{0.3242}  & 0.0000          & \textit{0.9576}   & 0.0002           \\
\textbf{F9}                                                                    & 1.0000            & 0.0000            & 1.0000              & 0.0000             & 0.7136            & 0.0000           & 0.8364           & 0.0094          & 0.4394           & 0.0017          & 1.0000            & 0.0002           \\
\textbf{Max}                                                                   & 1.0000            & 0.0315            & 1.0000              & 0.0270             & 1.0000            & 0.1050           & 1.0000           & 0.1295          & 1.0000           & 0.1448          & 1.0000            & \textbf{0.0052}  \\
\textbf{Min}                                                                   & 0.5667            & 0.0000            & 0.7985              & 0.0000             & 0.1500            & 0.0000           & 0.1530           & 0.0000          & 0.3242           & 0.0000          & \textbf{0.9576}   & 0.0000           \\ \hline
\textbf{Avg}                                                                   & 0.9305            & 0.0082            & 0.9574              & 0.0051             & 0.8580            & 0.0160           & 0.8535           & 0.0166          & 0.8677           & 0.0147          & \textbf{0.9921}   & \textbf{0.0010}  \\ \hline
\end{tabular}
\end{adjustbox}
\end{table}

\begin{figure}[!ht]
\centerline{\includegraphics[width=1.\columnwidth,height=0.4933\columnwidth]{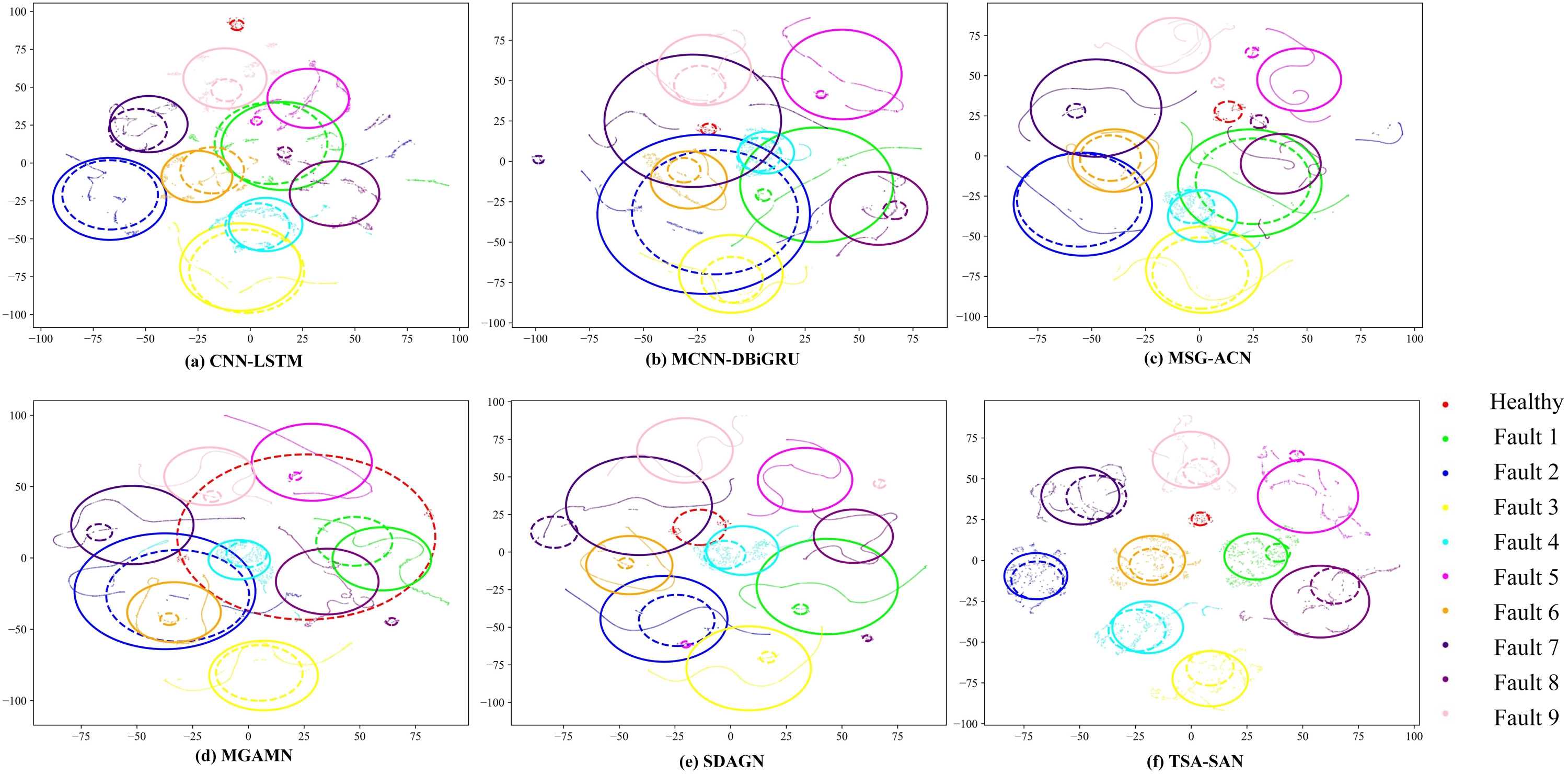}}
	\caption{Feature map visualization on CSTR dataset.}
	\label{Fig7}
\end{figure}

To evaluate the feature representation ability of different models, t-SNE is applied to visualize the features extracted by each model in task T6, as shown in \hyperref[Fig7]{ Fig. 7}. 
Different colors indicate different health state categories. 
To visually compare the differences in feature representations between seen and unseen distribution samples, we visualized the test samples using different line styles: samples consistent with the training distribution are marked with dashed outlines, while those from unseen distributions are marked with solid outlines. 
It is obvious that the proposed model achieved clear inter-class separation and improved intra-class compactness.
The data distributions of the same health state category under different domains are effectively aligned.
These results validate the proposed model's strong generalization capability under heterogeneous domains.

\subsubsection{TE process}
The accuracy results of all methods on TE dataset are summarized in \hyperref[Table9]{Table 9}.
The proposed model TSA-SAN consistently achieved the highest diagnostic accuracy of over 95\% in all four tasks.
In contrast, all other state-of-the-art models have accuracies below 85\%, which indicates a significant performance gap between the comparison models and the proposed model.
In terms of average accuracy, TSA-SAN outperformed all the comparison models with improvements of 61.03\%, 61.12\%, 28.21\%, 50.67\%,  and 53.87\% over CNN-LSTM, MCNN-DBiGRU, MAG-ACN, MGAMN, and SDAGN respectively.
It is worth noting that task T1 is regarded as the most challenging task, and all of the models got the lower accuracy on this task.
Nevertheless, the proposed TSA-SAN model still achieved a classification accuracy of 95.17\%, which is 53.48\% higher than the second-ranked MSG-ACN method. 
This significant improvement highlights the superior effectiveness of TSA-SAN in heterogeneous domain scenarios.
Compared with CNN-LSTM and MCNN-DBiGRU, which are designed for fault diagnosis under a single operating mode, the sample generation-based models such as MSG-ACN, MGAMN, and SDAGN all showed higher average accuracy. 
This suggests that generating samples based on heterogeneous domains does enhance the generalization ability to some extent.
However, these improvements are still not as good as those achieved by the proposed TSA-SAN.
This is primarily attributed to the large distributional divergence between different operating modes, which makes it difficult for existing models to generate samples that accurately reflect the real distributions. 
In contrast, TSA-SAN explicitly aligns the data distributions from different modes in the shared space and employs interpolation-based sample synthesis, thereby improving the model's adaptability to various data distributions. 

\begin{table}[!ht]
	\centering
	\caption{Accuracy results on TE process dataset.}
	\label{Table9}
\begin{tabular}{lllll|l}
\hline
\textbf{Model}       & \textbf{T1}      & \textbf{T2}      & \textbf{T3}      & \textbf{T4}      & \textbf{Avg}     \\ \hline
\textbf{CNN-LSTM}    & 14.61\%          & 51.11\%          & 50.13\%          & 27.41\%          & 35.82\%          \\
\textbf{MCNN-DBiGRU} & 15.79\%          & 62.50\%          & 30.37\%          & 34.25\%          & 35.73\%          \\
\textbf{MSG-ACN}     & 41.69\%          & 83.82\%          & 70.20\%          & 78.83\%          & 68.64\%          \\
\textbf{MGAMN}       & 28.19\%          & 67.87\%          & 42.09\%          & 46.56\%          & 46.18\%          \\
\textbf{SDAGN}       & 26.72\%          & 63.66\%          & 39.44\%          & 42.11\%          & 42.98\%          \\
\textbf{TSA-SAN}     & \textbf{95.17\%} & \textbf{98.10\%} & \textbf{97.36\%} & \textbf{96.76\%} & \textbf{96.85\%} \\ \hline
\end{tabular}
\end{table}

\begin{figure}[!ht]
\centerline{\includegraphics[width=1.\columnwidth,height=0.483\columnwidth]{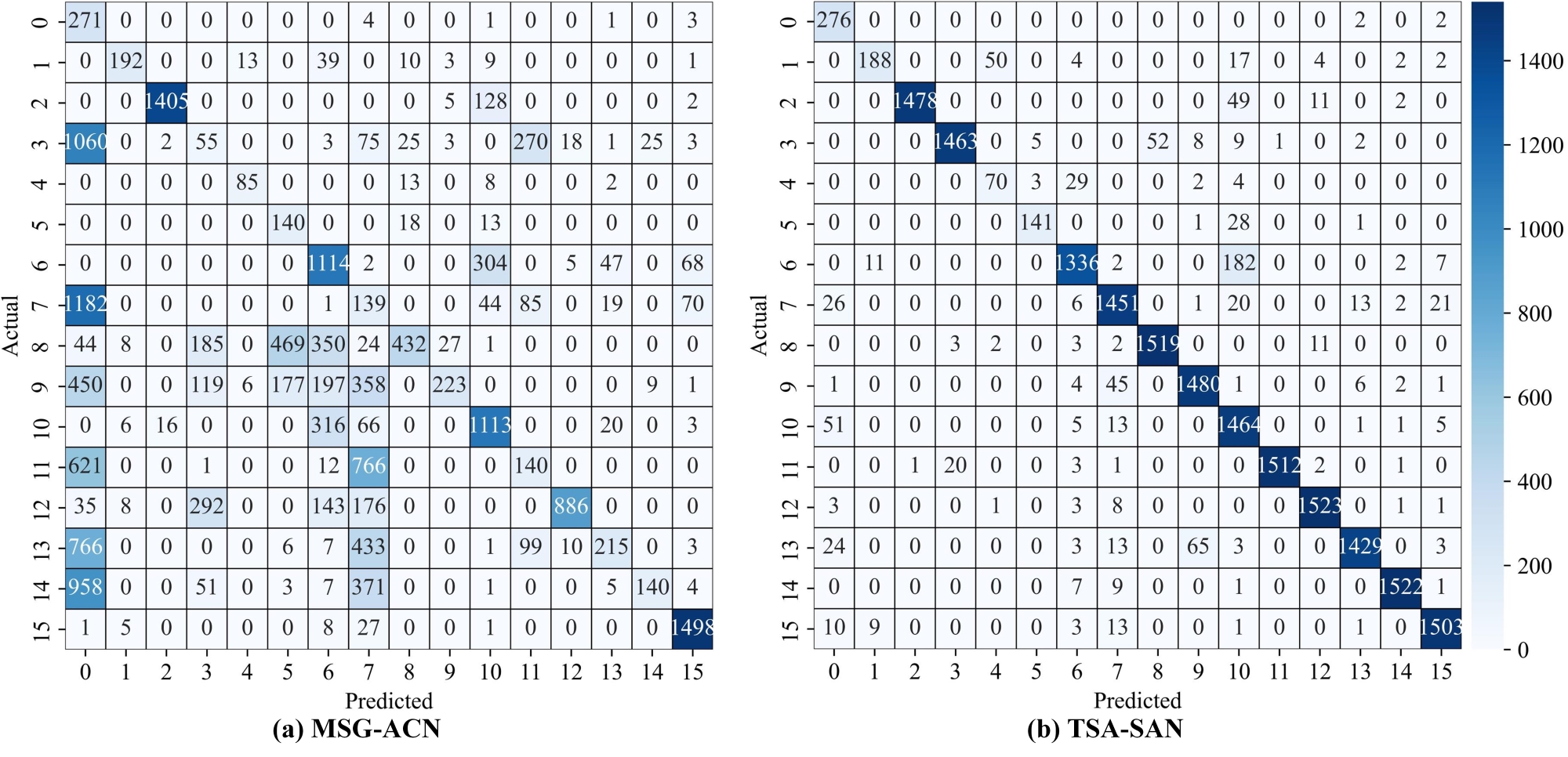}}
	\caption{Confusion matrix comparison.}
	\label{Fig8}
\end{figure}

Using task T1 as an example, the proposed TSA-SAN model is compared with the best performing comparison model MSG-ACN, which achieves the second highest accuracy in \hyperref[Table9]{Table 9}.
The confusion matrices of the two models are shown in \hyperref[Fig8]{ Fig. 8}. 
The MSG-ACN model clearly misclassified a large number of samples from fault categories F4 (label 3), F10 (label 7), F12 (label 9), F14 (label 11), F18 (label 13) and F19 (label 14) as healthy category N (label 0).
In contrast, the proposed TSA-SAN model effectively separates samples from different health state categories.
TSA-SAN got the largest number of correctly predicted samples along the diagonal, which indicates its superior diagnosis performance under heterogeneous domain scenarios.

Using task T2 as an example, the comparison results of the FDR and FPR scores of different models are listed in \hyperref[Table10]{Table 10}. 
The proposed TSA-SAN model significantly outperforms all comparison models in terms of average FDR scores, while its average FPR is reduced by nearly an order of magnitude.
Additionally, the proposed model attained a minimum FDR of 0.9383, which is obviously higher than those of the comparison models, and a maximum FPR of only 0.0052, which is the lowest among all models.  
These results indicate that the proposed TSA-SAN model exhibits superior consistency and stability in classifying different health state categories.

\begin{table}[!ht]
	\caption{FDR and FPR results on TE dataset.}
	\label{Table10}
            \begin{adjustbox}{width=1.4\textwidth,center=\textwidth}
\begin{tabular}{lllllllllllll}
\hline
\multirow{2}{*}{\textbf{\begin{tabular}[c]{@{}l@{}}Fault \\ No.\end{tabular}}} & \multicolumn{2}{c}{\textbf{CNN-LSTM}} & \multicolumn{2}{c}{\textbf{MCNN-DBiGRU}} & \multicolumn{2}{c}{\textbf{MSG-ACN}} & \multicolumn{2}{c}{\textbf{MGAMN}} & \multicolumn{2}{c}{\textbf{SDAGN}} & \multicolumn{2}{c}{\textbf{SATSFN}} \\ \cline{2-13} 
                                                                               & \textbf{FDR}      & \textbf{FPR}      & \textbf{FDR}        & \textbf{FPR}       & \textbf{FDR}      & \textbf{FPR}     & \textbf{FDR}     & \textbf{FPR}    & \textbf{FDR}     & \textbf{FPR}    & \textbf{FDR}     & \textbf{FPR}     \\ \hline
\textbf{N}                                                                     & 0.9893            & 0.0944            & 0.9964              & 0.1017             & 0.9714            & 0.0488           & 0.9929           & \textit{0.1205} & 0.9750           & \textit{0.1268} & 0.9786           & \textit{0.0051}  \\
\textbf{F1}                                                                    & 0.9539            & 0.0019            & 0.9838              & 0.0002             & 0.9883            & 0.0038           & 0.9753           & 0.0000          & 0.9812           & 0.0011          & 0.9935           & 0.0033           \\
\textbf{F2}                                                                    & 0.9623            & 0.0175            & 0.9396              & 0.0145             & 0.9701            & 0.0023           & 0.8896           & 0.0023          & 0.9552           & 0.0022          & 0.9792           & 0.0003           \\
\textbf{F4}                                                                    & 0.2922            & \textit{0.1210}   & \textit{0.0909}     & 0.0013             & \textit{0.0909}   & 0.0025           & 0.9422           & 0.0066          & 0.0909           & 0.0032          & 0.9851           & 0.0003           \\
\textbf{F6}                                                                    & 0.9474            & 0.0003            & 0.7970              & 0.0005             & 1.0000            & 0.0001           & 0.1955           & 0.0000          & 0.9474           & 0.0000          & 1.0000           & 0.0001           \\
\textbf{F7}                                                                    & 0.9688            & 0.0025            & 0.9818              & 0.0001             & 1.0000            & 0.0001           & 0.5208           & 0.0062          & 0.9942           & 0.0007          & 1.0000           & 0.0002           \\
\textbf{F8}                                                                    & 0.8929            & 0.0247            & 0.7844              & 0.0032             & 0.8597            & 0.0029           & 0.6143           & 0.0497          & 0.7396           & 0.0252          & \textit{0.9383}  & 0.0024           \\
\textbf{F10}                                                                   & 0.0909            & 0.1056            & 0.0909              & 0.0536             & 0.6506            & 0.0007           & 0.4786           & 0.0600          & \textit{0.0799}  & 0.0926          & 0.9857           & 0.0017           \\
\textbf{F11}                                                                   & 0.0909            & 0.0281            & 0.5942              & \textit{0.1138}    & 0.9545            & \textit{0.0684}  & 0.8766           & 0.0077          & 0.6519           & 0.0688          & 0.9896           & 0.0015           \\
\textbf{F12}                                                                   & \textit{0.0903}   & 0.0120            & 0.0987              & 0.0023             & 0.8753            & 0.0099           & \textit{0.1468}  & 0.0073          & 0.0883           & 0.0114          & 0.9903           & 0.0003           \\
\textbf{F13}                                                                   & 0.8922            & 0.0753            & 0.9610              & 0.0296             & 0.9481            & 0.0081           & 0.7253           & 0.0314          & 0.8987           & 0.0329          & 0.9578           & 0.0034           \\
\textbf{F14}                                                                   & 0.1896            & 0.0234            & 0.7461              & 0.0115             & 0.9968            & 0.0000           & 0.5006           & 0.0001          & 0.7773           & 0.0000          & 0.9974           & 0.0000           \\
\textbf{F17}                                                                   & 0.9331            & 0.0000            & 0.9838              & 0.0002             & 0.9877            & 0.0000           & 0.9052           & 0.0000          & 0.9740           & 0.0000          & 0.9922           & 0.0002           \\
\textbf{F18}                                                                   & 0.4331            & 0.0030            & 0.2935              & 0.0070             & 0.9461            & 0.0032           & 0.8903           & 0.0027          & 0.5260           & 0.0066          & 0.9539           & 0.0004           \\
\textbf{F19}                                                                   & 0.0909            & 0.0091            & 0.1403              & 0.0006             & 0.4500            & 0.0198           & 0.5487           & 0.0307          & 0.0909           & 0.0073          & 0.9909           & 0.0000           \\
\textbf{F20}                                                                   & 0.1494            & 0.0012            & 0.9786              & 0.0569             & 0.9786            & 0.0003           & 0.4714           & 0.0129          & 0.9753           & 0.0041          & 0.9786           & 0.0007           \\
\textbf{Max}                                                                   & 0.9893            & 0.1210            & 0.9964              & 0.1138             & 1.0000            & 0.0684           & 0.9929           & 0.1205          & 0.9942           & 0.1268          & 1.0000           & \textbf{0.0051}  \\
\textbf{Min}                                                                   & 0.0903            & 0.0000            & 0.0909              & 0.0001             & 0.0909            & 0.0000           & 0.1468           & 0.0000          & 0.0799           & 0.0000          & \textbf{0.9383}  & 0.0000           \\
\textbf{Avg}                                                                   & 0.5605            & 0.0325            & 0.6538              & 0.0248             & 0.8543            & 0.0107           & 0.6671           & 0.0211          & 0.6716           & 0.0239          & \textbf{0.9819}  & \textbf{0.0012}  \\ \hline
\end{tabular}
\end{adjustbox}
\end{table}

\begin{figure}[!ht]
\centerline{\includegraphics[width=1.\columnwidth,height=0.502\columnwidth]{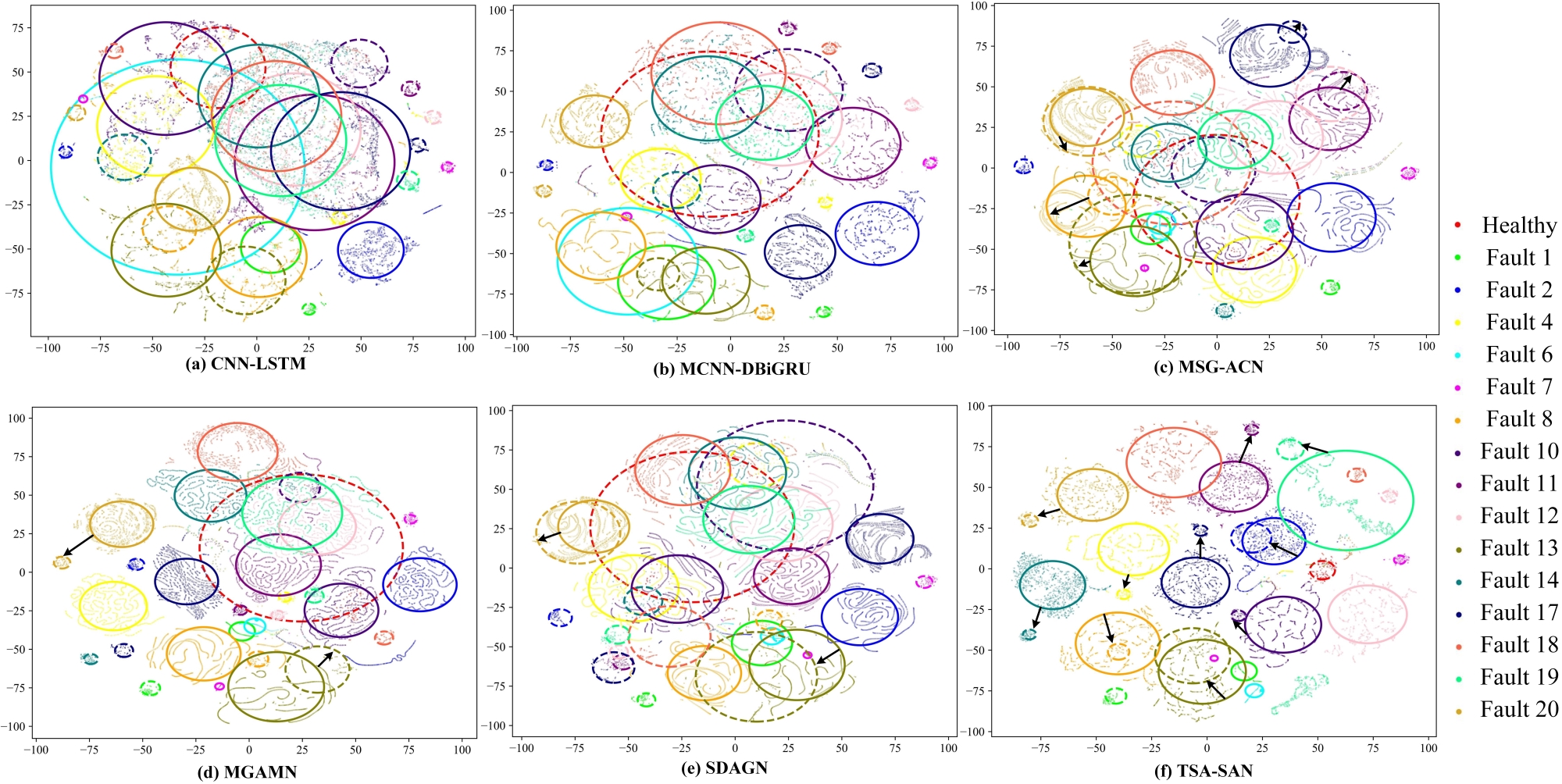}}
	\caption{Feature map visualization on TE dataset.}
	\label{Fig9}
\end{figure}

Similarly, the feature visualization results on the TE dataset are presented in \hyperref[Fig9]{ Fig. 9}. 
The samples from seen and unseen distributions are also outlined using dashed and solid contours, respectively.
To highlight the effectiveness of feature alignment, arrows are drawn between the contours of seen and unseen distributions belonging to the same category when they appear visually close in the feature space. 
The visualization results demonstrate two key strengths of the proposed model in feature learning. 
First, the features of different categories are well separated with minimal overlap in the embedding space. 
Second, the samples belonging to the same category but originating from seen and unseen distributions are close together in the feature space.
These results further demonstrate that the proposed model exhibits strong discriminative ability on seen data distributions and achieves superior generalization performance on unseen distributions.

\subsection{Ablation study}
Comprehensive ablation studies were conducted to evaluate the contribution of each component, including distribution alignment sample generation strategy (DASG), interpolation-based sample synthesis strategy (ISS), self-adaptive instance normalization (SAIN) and temporal-spatial attention mechanism (TSAM). 
The experiment settings and results are summarized in \hyperref[Table11]{Table 11} and \hyperref[Table12]{Table 12}, respectively.
Obviously, DASG has the greatest influence on model performance. 
Removing DASG led to a 39.29\% decrease in average diagnostic accuracy, which highlights the critical role of generating effective samples in enhancing the model's generalization ability.
The TSAM also contributed significantly, with its removal resulting in a 7.6\% reduction in average accuracy. 
This suggests that it is critical to focus on discriminative temporal-spatial regions in complex data distributions.
Compared with A2, the introduction of the ISS in proposed model improves the average diagnostic accuracy by 6.38\%. 
This may be attributed to the fact that the transition process between healthy and fault states is also considered as the fault samples, which enhances the model's ability to classify early faults and facilitates the learning of more generalized fault representations.
The main difference between A3 and A4 lies in the use of normalization.
A3 does not employ normalization, while A4 introduces the instance normalization. 
Compared to the full TSA-SAN model, their average accuracies decreased by 5.15\% and 5.07\%, respectively. 
This result demonstrates the effectiveness of SAIN in suppressing domain-related interference while retaining the necessary discriminative features.
Overall, the complete model integrating DASG, ISS, SAIN, and TSAM achieved the best diagnostic performance in all four tasks, which indicates that each component is indispensable. 
The combination of sample generation and self-adaptive temporal-spatial feature extraction provides an effective solution to the fault diagnosis across heterogeneous domains.

\begin{table}[!ht]
	\centering
	\caption{Ablation study settings.}
	\label{Table11}
\begin{tabular}{llllll}
\hline
\textbf{Component} & \textbf{A1}     & \textbf{A2}      & \textbf{A3}   & \textbf{A4}  & \textbf{A5}  \\ \hline
\textbf{DASG}       &                 & $\checkmark$     & $\checkmark$  & $\checkmark$ & $\checkmark$ \\
\textbf{ISS}       & \multicolumn{2}{l}{$\checkmark$}   & $\checkmark$  & $\checkmark$ & $\checkmark$ \\
\textbf{SAIN}      & $\checkmark$    & \multicolumn{2}{l}{$\checkmark$} & IN           & $\checkmark$ \\
\textbf{TSAM}      & $\checkmark$    & $\checkmark$     & $\checkmark$  & $\checkmark$ &              \\ \hline
\end{tabular}
\end{table}

\begin{table}[!ht]
	\centering
	\caption{Accuracy results of ablation experiments on the TE dataset.}
	\label{Table12}
\begin{tabular}{lllll|l}
\hline
\textbf{Model} & \textbf{T1} & \textbf{T2} & \textbf{T3} & \textbf{T4} & \textbf{Avg} \\ \hline
\textbf{A1}       & 24.29\% & 80.87\% & 60.28\% & 64.79\% & 57.56\% \\
\textbf{A2}       & 75.13\% & 97.88\% & 95.33\% & 93.52\% & 90.47\% \\
\textbf{A3}       & 78.29\% & 96.93\% & 95.27\% & 96.30\% & 91.70\% \\
\textbf{A4}       & 84.54\% & 97.09\% & 90.54\% & 94.96\% & 91.78\% \\
\textbf{A5}       & 86.34\% & 95.80\% & 82.82\% & 92.02\% & 89.25\% \\
\textbf{TSA-SAN}  & \textbf{95.17\%} & \textbf{98.10\%} & \textbf{97.36\%} & \textbf{96.76\%} & \textbf{96.85\%} \\ \hline
\end{tabular}
\end{table}

To further validate the advantages of the proposed model, the comparison models were trained on the dataset augmented using the proposed data generation strategy. 
The diagnostic results are summarized in \hyperref[Table13]{Table 13}. 
Compared to the results in \hyperref[Table9]{Table 9}, all comparison models exhibit significant improvements in accuracy, which confirms the effectiveness of the sample generation strategy.
Under this setting, the proposed model achieves the highest diagnostic accuracy in three tasks, except for task T2, where its performance is only 0.05\% lower than that of MCNN-DBiGRU. 
Additionally, it outperforms CNN-LSTM, MCNN-DBiGRU, MSG-ACN, MGAMN, and SDAGN by 4.67\%, 5.58\%, 7.96\%, 6.46\%, 12.35\% and 8.11\% in average accuracy, respectively.
These experiment results further verify the advantages of the proposed model in learning discriminative and generalizable feature representations.

\begin{table}[!ht]
	\centering
	\caption{Accuracy results of further study on TE process dataset.}
	\label{Table13}
\begin{tabular}{lllll|l}
\hline
\textbf{Model}       & \textbf{T1}      & \textbf{T2}      & \textbf{T3}      & \textbf{T4}      & \textbf{Avg}     \\ \hline
\textbf{CNN-LSTM}    & 82.80\%          & 96.37\%          & 94.01\%          & 95.52\%          & 92.18\%          \\
\textbf{MCNN-DBiGRU} & 76.68\%          & \textbf{98.15\%} & 93.54\%          & 96.71\%          & 91.27\%          \\
\textbf{MSG-ACN}     & 76.68\%          & 95.86\%          & 89.01\%          & 94.02\%          & 88.89\%          \\
\textbf{MGAMN}       & 79.24\%          & 95.81\%          & 94.00\%          & 92.52\%          & 90.39\%          \\
\textbf{SDAGN}       & 67.45\%          & 95.83\%          & 80.97\%          & 93.73\%          & 84.50\%          \\
\textbf{TSA-SAN}     & \textbf{95.17\%} & 98.10\%          & \textbf{97.36\%} & \textbf{96.76\%} & \textbf{96.85\%} \\ \hline
\end{tabular}
\end{table}

\subsection{Case study on practical industrial system}
\begin{figure}[!ht]
\centerline{\includegraphics[width=1.\columnwidth,height=0.538\columnwidth]{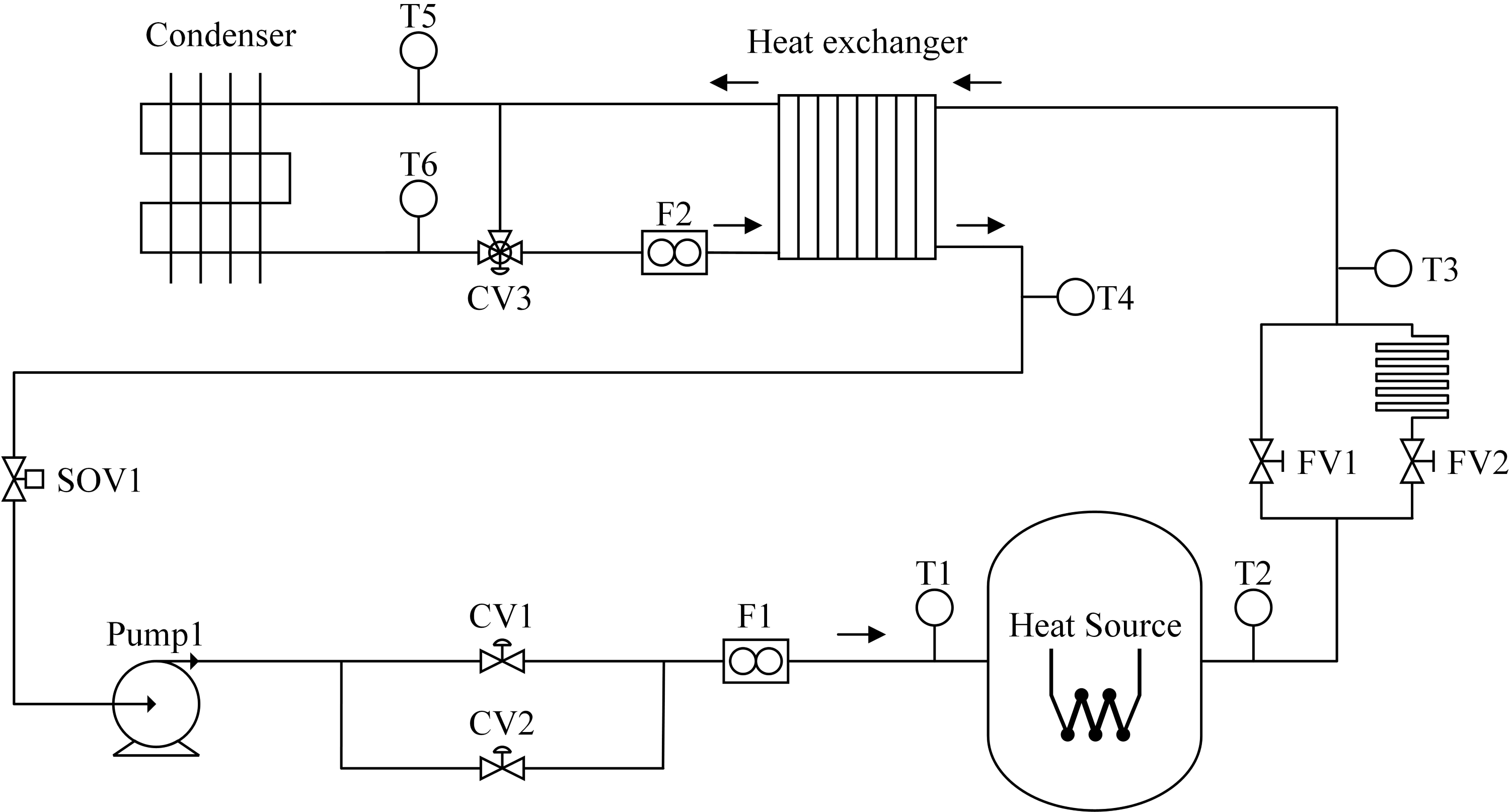}}
	\caption{Structure of IPCTF.}
	\label{Fig10}
\end{figure}

The intelligent process control-test facility (IPCTF) developed by Wuhan University of Technology is an experimental platform for industrial process control. 
In this system, the temperature regulation of the hot loop is achieved by the coupled cooling loop.
Nine critical variables were selected in this study, including six temperature measurements (T1-T6), two flow measurements (F1,F2), and one heat source power measurement, as shown in \hyperref[Fig10]{Fig. 10}. 
The operating modes are defined by changing the setpoint of the cooling loop and the power of the heat source. 
Two typical faults were introduced: pump blockage in the hot loop and pipeline blockage in the cooling loop. 
The experimental configurations are detailed in \hyperref[Table14]{Table 14}.

\begin{table}[!ht]
\centering
	\label{Table14}
		\caption{Task settings of IPCTF used in HDFD experiments.}
\begin{threeparttable}
\begin{tabular}{llll}
\hline
\textbf{Mode No.}                          & \multicolumn{2}{l}{\textbf{Power of the heat source (kW)}}                                  & \textbf{Setpoint of the cooling loop (℃)}        \\ \hline
\textbf{M1}                                & \multicolumn{2}{l}{10}                                                                                                                    & 29                                                                                             \\
\textbf{M2}                                & \multicolumn{2}{l}{8}                                                                                                                     & 24                                                                                             \\ \hline
\multicolumn{2}{l}{\textbf{Training set}}                                                                                 & \multicolumn{2}{l}{\textbf{Test set}}                                                                                                                       \\ \hline
\multicolumn{2}{l}{\begin{tabular}[c]{@{}l@{}}M1: normal, pump blockage   \\ M2: normal, pipeline blockage\end{tabular}} & \multicolumn{2}{l}{\begin{tabular}[c]{@{}l@{}}M1: normal, pump blockage, pipeline blockage   \\ M2: normal, pump blockage, pipeline blockage\end{tabular}} \\ \hline
\end{tabular}
\end{threeparttable}
\end{table}

\begin{figure}[!ht]
\centerline{\includegraphics[width=0.5\columnwidth,height=0.4178\columnwidth]{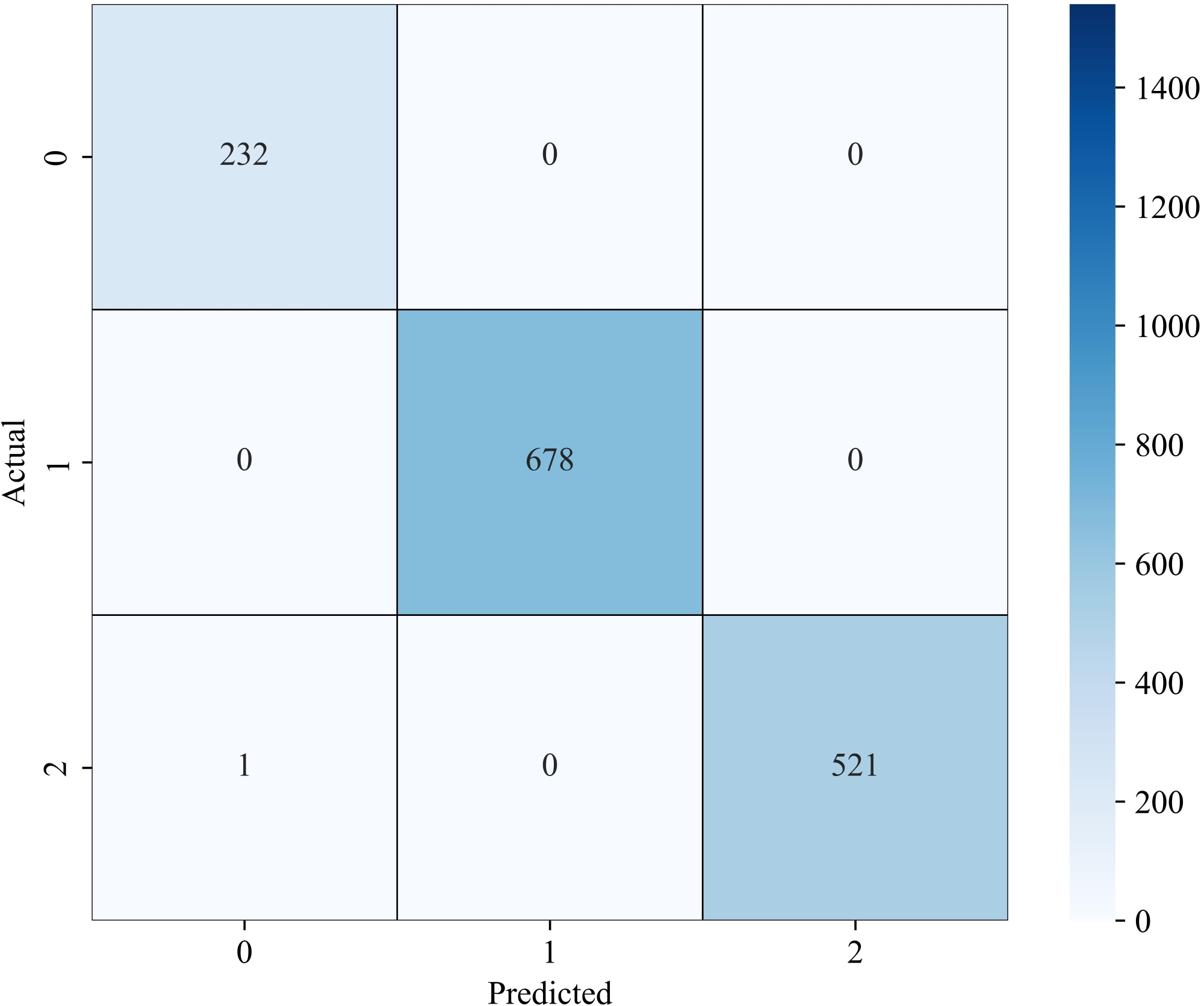}}
	\caption{Confusion matrix of the proposed method on IPCTF dataset.}
	\label{Fig11}
\end{figure}

The proposed TSA-SAN model was applied to the IPCTF process and the confusion matrix obtained is shown in \hyperref[Fig11]{Fig. 11}. 
The result shows that only one sample from the pipeline blockage was misclassified as a healthy category.
The proposed TSA-SAN model achieved an overall accuracy of 99.93\%. 
These results demonstrate that the proposed model also exhibits excellent performance for fault diagnosis across heterogeneous domains tasks in practical industrial systems.

\section{Conclusion}
\label{sec5}
This paper proposes a novel self-adaptive temporal-spatial attention network (TSA-SAN) for fault diagnosis under heterogeneous domain scenarios. 
This model enhances the generalization performance through sample generation and self-adaptive temporal-spatial feature extraction. 
Extensive experiments on two public benchmark datasets and one real industrial dataset demonstrate that the proposed method is noticeably superior to the existing state-of-the-art methods. 
The key conclusions are summarized as follows.

\begin{itemize}
\item The distribution alignment sample generation strategy effectively realizes cross-domain fault sample enhancement by learned inter-domain mapping.

\item The interpolation-based sample synthesis strategy generates diverse fault samples and improves the adaptability of the model to a wide range of data distributions.

\item The self-adaptive instance normalization suppresses irrelevant information while preserving discriminative features.

\item The temporal-spatial attention mechanism strengthens the model's ability to capture critical features in both temporal and spatial dimensions.
\end{itemize}

Despite the promising results, this study still has certain limitations. It does not explicitly consider the potentially complex or dynamic mappings between different domains. 
Future work could further focus on developing robust diagnostic models under complex or time-varying operating conditions. 
In addition, the identification of unknown faults is also an important direction for further exploration.


\bibliographystyle{elsarticle-num} 
\bibliography{ref}
\end{document}